\documentclass[lettersize,journal]{IEEEtran}
\usepackage{amsmath,amsfonts}
\usepackage{algorithmic}
\usepackage{algorithm}
\usepackage{array}
\usepackage[caption=false,font=normalsize,labelfont=sf,textfont=sf]{subfig}
\usepackage{textcomp}
\usepackage{stfloats}
\usepackage{url}
\usepackage{verbatim}
\usepackage{graphicx}
\usepackage{cite}
\usepackage{orcidlink}
\usepackage{amsthm,amssymb}
\usepackage{mathrsfs}
\usepackage{multicol}
\usepackage{multirow}
\usepackage{soul}
\usepackage{color}
\usepackage{rotating}
\usepackage{tablefootnote}
\usepackage[flushleft]{threeparttable}
\usepackage[switch]{lineno} 
\begin{document}

\title{Toward Automated Algorithm Design:\\ A Survey and Practical Guide to Meta-Black-Box-Optimization}


\author{Zeyuan Ma~\orcidlink{0000-0001-6216-9379}, Hongshu Guo~\orcidlink{0000-0001-8063-8984},  Yue-Jiao Gong~\orcidlink{0000-0002-5648-1160},~\IEEEmembership{Senior Member,~IEEE} \\ Jun Zhang~\orcidlink{0000-0001-7835-9871},~\IEEEmembership{Fellow,~IEEE}
and Kay Chen TAN~\orcidlink{0000-0002-6802-2463},~\IEEEmembership{Fellow,~IEEE} 
\thanks{Zeyuan Ma, Hongshu Guo, and Yue-Jiao Gong are with the School of Computer Science and Engineering, South China University of Technology, Guangzhou 510006, China (E-mail: gongyuejiao@gmail.com)}
\thanks{Jun Zhang is with Nankai University, Tianjin, China and Hanyang University, Seoul, South Korea. (E-mail: junzhang@ieee.org).}
\thanks{Kay Chen TAN is with Hong Kong Polytechnic University, Hong Kong, China. (E-mail: kaychen.tan@polyu.edu.hk).}
\thanks{Corresponding author: Yue-Jiao Gong}
}



\maketitle

\begin{abstract}
In this survey, we introduce Meta-Black-Box-Optimization~(MetaBBO) as an emerging avenue within the Evolutionary Computation~(EC) community, which incorporates Meta-learning approaches to assist automated algorithm design.  Despite the success of MetaBBO, the current literature provides insufficient summaries of its key aspects and lacks practical guidance for implementation. 
To bridge this gap, we offer a comprehensive review of recent advances in MetaBBO, providing an in-depth examination of its key developments. We begin with a unified definition of the MetaBBO paradigm, followed by a systematic taxonomy of various algorithm design tasks, including algorithm selection, algorithm configuration, solution manipulation, and algorithm generation. Further, we conceptually summarize different learning methodologies behind current MetaBBO works, including reinforcement learning, supervised learning, neuroevolution, and in-context learning with Large Language Models. A comprehensive evaluation of the latest representative MetaBBO methods is then carried out, alongside an experimental analysis of their optimization performance, computational efficiency, and generalization ability. Based on the evaluation results, we meticulously identify a set of core designs that enhance the generalization and learning effectiveness of MetaBBO. Finally, we outline the vision for the field by providing insight into the latest trends and potential future directions. Relevant literature will be continuously collected and updated at 
\url{https://github.com/MetaEvo/Awesome-MetaBBO}.
\end{abstract}


\section{Introduction}

Optimization techniques have been central to research for decades~\cite{2005ec-survey,2019opt-survey}, with methods applied across engineering~\cite{2016opt-engi-survey}, economics~\cite{2012opt-economics-survey}, and science~\cite{2018opt-discovery}. The optimization problems can be classified into White-Box~\cite{2006hinton} and Black-Box~\cite{2017ecrl} types. White-Box problems, with transparent structures, allow efficient optimization using gradient-based algorithms like SGD~\cite{2016sgd}, Adam~\cite{2014adam}, and BFGS~\cite{1989bfgs}. In contrast, Black-Box Optimization (BBO) only provides objective values for solutions, making the analysis and search of the problem space even more challenging.

Evolutionary Computation (EC), including Evolutionary Algorithms (EAs) and Swarm Intelligence (SI), is widely recognized as an effective gradient-free approach for solving BBO problems~\cite{2016ea-survey}. Over the past decades, EC methods have been extensively applied to various optimization challenges~\cite{cec2014,2008mop-benchmark,2011mmop,2007lsgo-benchmark,2021mtop}, due to their simplicity and versatility. Though effective for solving BBO problems, traditional EC is constrained by the no-free-lunch theorem~\cite{1997nofreelunch}, which asserts that no optimization algorithm can universally outperform others across all problem types, leading to performance trade-offs depending on the problem's characteristics. In response, various human-crafted methods have been developed including offline hyper-parameter optimization~\cite{hpo-1,hpo-2,hpo-3}, hyper-heuristics~\cite{HH-2000,HH-generation-app-1,HH-generation-app-2} and (self-)adaptive EC variants~\cite{1994aga,2009jade,2009apso,2015glpso,hansen2016cmaes,2013shade,2014lshade,2022nl-shade-lbc}. However, they face several limitations. 1)~Limited generalization: these methods often focus on a specific set of problems, limiting their generalization due to customized designs.
2)~Labor-intensive: designing adaptive mechanisms requires both deep knowledge of EC domain and the target optimization problem, making it a complex task.
3)~Additional parameters: many adaptive mechanisms introduce extra hyper-parameters, which can significantly impact performance.
4)~Sub-optimal performance: despite increased efforts, design biases and delays in reactive adjustments often lead to sub-optimal outcomes.
Given this, a natural question arises: can we automatically design effective BBO algorithms while minimizing the dependence on expert input? A recently emerging research topic, known as Meta-Black-Box-Optimization~(MetaBBO)~\cite{2024metabox}, has shown possibility of leveraging the generalization strength of Meta-learning~\cite{2019metalearning} to enhance the optimization performance of BBO algorithms in the minimal expertise cost. MetaBBO follows a bi-level paradigm: the meta level typically maintains a policy that takes the low-level optimization information as input and then automatically dictates desired algorithm design for the low-level BBO optimizer. The low-level BBO process evaluates the suggested algorithm design and returns a feedback signal to the meta-level policy regarding the performance gain. The meta-objective of MetaBBO is to meta-learn a policy that maximizes the performance of the low-level BBO process, over a problem distribution. Once the training completes, the learned meta-level policy can be directly applied to address unseen optimization problems, hence reducing the need for expert knowledge to adapt BBO algorithms. 

Numerous valuable ideas have been proposed and discussed in existing MetaBBO research. From the perspective of algorithm design tasks~(meta tasks) that the meta-level policy can address, 
those MetaBBO works can be categorized into four branches: 1)~Algorithm Selection, where for solving the given problem, a proper BBO algorithm is selected by the meta-level policy from a pre-collected optimizer/operator pool. 2)~Algorithm Configuration, where the hyper-parameters and/or operators of a BBO algorithm are adjusted by the meta-level policy to adapt for the given problem. 3)~Solution Manipulation, where the meta-level policy is trained to act as a BBO algorithm to manipulate and evolve solutions. 4)~Algorithm Generation, where each algorithmic component and the overall workflow are generated by the meta-level policy as a novel BBO algorithm. From the perspective of learning paradigms adopted for training the meta-level policy, different learning methods such as reinforcement learning~(MetaBBO-RL)~\cite{2019deddqn,2022rlhpsde,2020qlpso,sun2021lde,ma2024gleet,2021dedqn,2024rldas}, auto-regressive supervised learning~(MetaBBO-SL)~\cite{2024llamoco,2024evotf,2024ribbo,2017rnnoi,2019rnnopt,2024glhf}, neuroevolution~(MetaBBO-NE)~\cite{2021ltopomdp,2023les,2023lga}, and Large Language Models (LLMs)-based in-context learning~(MetaBBO-ICL)~\cite{2023evoprompt,2024ribbo,2024evollm,liu2024eoh,ahmaditeshnizi2023optimus} have been investigated in existing works. From the perspective of low-level BBO process, MetaBBO has been instantiated to various optimization scenarios such as single-objective optimization~\cite{sun2021lde,ma2024gleet,2024rldas}, multi-objective optimization~\cite{tahernezhad2024r2-rlmoea,wang2024mrl-moea}, multi-modal optimization~\cite{lian2024rlemmo}, large scale global optimization~\cite{2023lga,2023les,2024glhf}, and multi-task optimization~\cite{2024meta-mto,2024mto-llm}. Such an intricate combination of algorithm design tasks, learning paradigms, and low-level BBO scenarios makes it challenging for new practitioners to systematically learn, use, and develop MetaBBO methods. Unfortunately, there is still a lack of a comprehensive survey and practical guide to the advancements in MetaBBO. 
\begin{figure*}[t]
    \centering
    \includegraphics[width=0.9\linewidth]{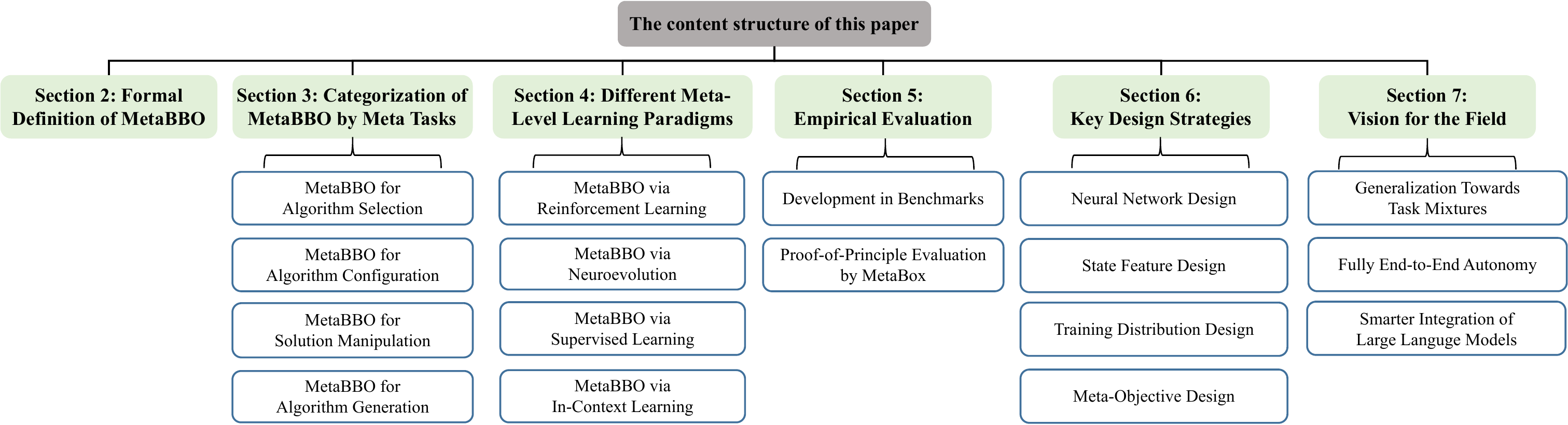}
    \caption{
    Roadmap of the content structure, beginning with a concept introduction, followed by a review of existing methods across different taxonomies, a evaluation of selected methods, and a summary of key design strategies and future vision.}
    \label{fig:roadmap}
\end{figure*}

While some related surveys discussed the integration of learning systems into EC algorithm designs, they have several limitations: 1)~Previous surveys~\cite{surver-add-1,surver-add-2,surver-add-3} focus on one or two algorithm design tasks, such as algorithm configuration~\cite{surver-add-3} and algorithm generation~\cite{surver-add-1,surver-add-2}. These surveys therefore show short in providing comprehensive review and comparison analysis on all four design tasks. 2)~Some surveys~\cite{2019rleassurvey,2023rleassurvey,2024rleassurvey,2024tkrleassurvey} focus on a particular learning paradigm - RL~\cite{sutton2018reinforcement}. 
However, in MetaBBO, various learning paradigms can be adopted, each with distinct characteristics. 
3)~In addition to reviewing relevant papers, existing surveys lack a practical guide that provides a comprehensive experimental evaluation of MetaBBO methods and a summary of key design strategies, falling short in offering in-depth evaluations or actionable insights for implementing MetaBBO methods.

\begin{figure}[t]
    \centering
    \includegraphics[width=0.7\columnwidth]{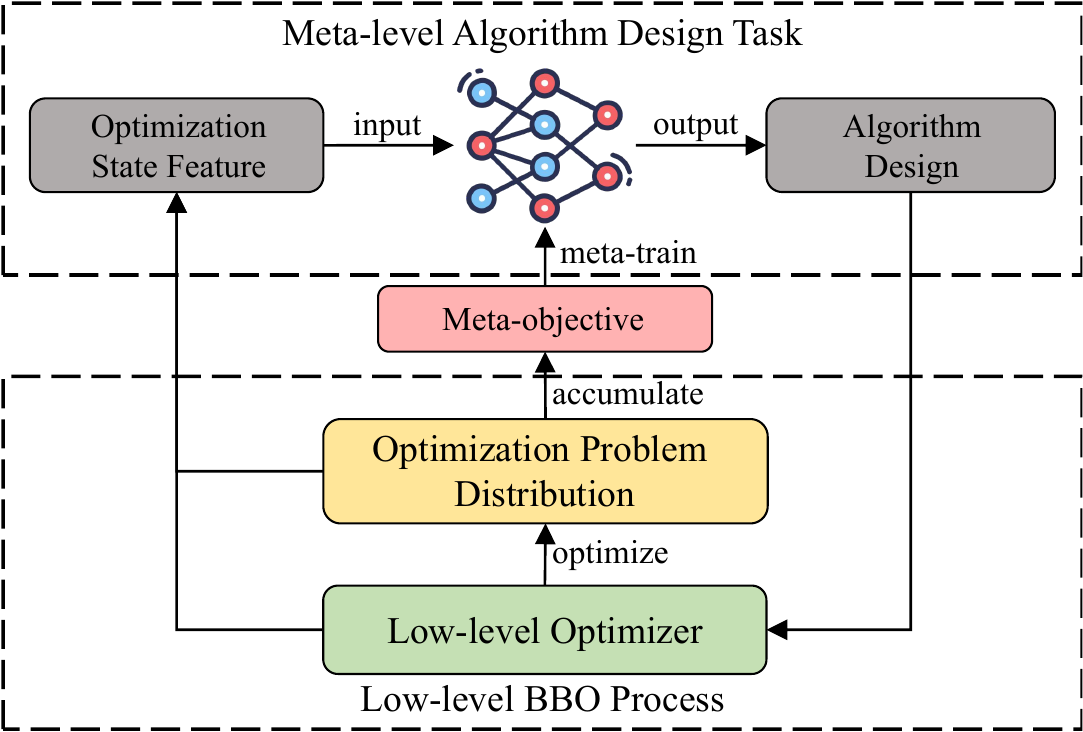}
    \caption{A conceptual overview of the bi-level learning framework of MetaBBO, illustrating the interactions between its core components to clarify the overall workflow.}
    \label{fig:metabbo}
\end{figure}

To address the gaps in previous surveys, this paper provides a more comprehensive coverage of the MetaBBO field. Fig.~\ref{fig:roadmap} offers a roadmap to help readers quickly navigate the overall content structure. We first provide a formal definition of MetaBBO in Section~\ref{sec:2}. Subsequently, we identify four main algorithm design tasks in existing MetaBBO works and their working scenarios in Section~\ref{sec:3}. In Section~\ref{sec:4}, we further elaborate four learning paradigms, with easy-to-follow technical details. Section~\ref{sec:5} provides a proof-of-principle performance evaluation on ten representative MetaBBO methods. According to the evaluation results, Section~\ref{sec:6} provides in-depth discussion about the key design strategies in MetaBBO. Finally, we outline the vision for the MetaBBO field in Section~\ref{sec:7}. The contributions of this survey are generally summarized as follows:
\begin{itemize}
    \item The first comprehensive survey that sorts out existing literature on MetaBBO.
    We provide a clear categorization of existing MetaBBO works according to four distinct meta-level tasks, along with a detailed elaboration of four different learning paradigms behind.
    \item A proof-of-principle evaluation is conducted to provide practical comparison between MetaBBO works, leading to an in-depth discussion over several key design strategies related to the learning effectiveness, training efficiency, and generalization. 
    \item In the end of this paper, we mark several interesting and promising future research directions of MetaBBO, focusing different aspects such as the generalization potential, the end-to-end workflow, and the integration of LLMs. 
\end{itemize}





\section{Definition of MetaBBO}\label{sec:2}
Meta-Black-Box-Optimization~(MetaBBO)~\cite{2024metabox} is derived from the Meta-learning paradigm~\cite{2019metalearning,mata-survey}. One of the roots of meta-learning could date back to 1987, where  Schmidhuber~\cite{meta-1} used Genetic Programming~(GP) as a learning method to learn better GP program in a self-referential way. Meta-learning is applicable to the learning of both models and algorithms~\cite{2017rnnoi}. For example, in~\cite{meta-2}, a prior knowledge-assisted learning method was proposed to learn synaptic learning
rule that is adaptable for diverse downstream learning tasks. In~\cite{2016meta-model}, a novel Recurrent Neural Network (RNN) model architecture is meta-learned and subsequently used as a classifier. In contrast, in ~\cite{2016meta-sgd}, a RNN model is meta-learned to serve as a gradient descent algorithm for optimizing other neural networks. The paradigm in ~\cite{2016meta-sgd} quickly becomes popular and the following works explore the possibility of such a paradigm in various white-box optimization scenarios ranging from first-order optimization~\cite{2016meta-sgd} to combinatorial optimization~\cite{2019AM}. To make a distinction with other applications of Learning to Learn, this research line is named by Learning to Optimize~(L2O). MetaBBO draws key inspiration from the L2O, while targeting black-box optimization scenarios. 
In this section, we provide an overview of the abstract workflow shared by existing MetaBBO methods, explaining the motivation of the core components in MetaBBO. MetaBBO operates within a bi-level framework, as depicted in Fig.~\ref{fig:metabbo}, and is detailed as follows.

We begin with the low-level BBO process. A key component at this level is the low-level optimizer $\mathcal{A}$. 
$\mathcal{A}$ represents a flexible concept, capable of being any off-the-shelf EC algorithm, its modern variants, an algorithm pool, or a structure for creating new algorithms (rather than a specific existing one). Another crucial element is the optimization problem distribution $\mathcal{P}$, representing a collection of optimization problem instances to be solved. Although the size of $\mathcal{P}$ could theoretically be infinite, facilitating Meta-learning on an infinite problem set is impossible. In practice, we instead sample a collection of $N$ instances $\{f_1, f_2, ...,f_N\}$ from $\mathcal{P}$ as the training set. 
A meta task $\mathcal{T}$ aims to automatically dictate an algorithm design $\omega \in \Omega$ for the low-level optimizer $\mathcal{A}$ for each problem instance in $\mathcal{P}$, where $\Omega$ denotes the algorithm design space of $\mathcal{A}$. For instance, in a basic DE optimizer~\cite{1997de}, its algorithm configuration~(e.g., values of the two hyper-parameters $F$ and $Cr$ that control the mutation and crossover strength) can be regarded as an algorithm design space $\Omega$. There are various algorithm design spaces, which are discussed in detail in Section~\ref{sec:3}. MetaBBO solves the meta task by learning a meta-level policy $\pi_\theta$ for the algorithm decision. 

Formally, for a meta-level algorithm design task $\mathcal{T}:=\{\mathcal{P},\mathcal{A},\Omega\}$, $\pi_\theta$ is trained to maximize the meta-objective $J(\theta)$:
\begin{align}\label{eq:1}
J(\theta) = \mathbb{E}_{f \in \mathcal{P}} \left[ \text{R}\left ( \mathcal{A}, \pi_\theta, f\right ) \right] &\approx \frac{1}{N}\sum_{i=1}^{N} \sum_{t=1}^{T} \text{perf}(\mathcal{A}, \omega_i^t, f_i) \notag\\
     \omega_i^t = \pi_\theta(s_i^t), & \quad s_i^t = \text{sf}(\mathcal{A},f_i,t)  
\end{align}
where $\text{sf}(\cdot)$ is a state feature extraction function, which captures the optimization state information from the interplay between the optimizer $\mathcal{A}$ and the problem instance $f_i$. The meta-level policy $\pi_\theta$ is parameterized by learnable parameters $\theta$. It receives $s_i^t$ as input and outputs an algorithm design $\omega_i^t$, which is then adopted by $\mathcal{A}$ to optimize $f_i$. A performance measurement function $\text{perf}(\cdot)$ is used to evaluate the performance gain obtained by this algorithm design decision. $\text{R}(\cdot)$ is accumulated performance gain during the low-level optimization of a problem instance. We approximate the meta-objective $J(\theta)$ as the average performance gain across a group of $N$ problem instances sampled from $\mathcal{P}$, over a certain number $T$ of optimization steps. To summarize, MetaBBO aims to search for an optimal meta-level policy $\pi_{\theta^*}$ which maximizes the meta-objective $J(\theta)$.

\textbf{MetaBBO vs. Hyper-Heuristics (HH):} It is worthy to note that MetaBBO closely aligns with the Hyper-Heuristics~(HH) paradigm introduced by Peter et al.~\cite{HH-2000} through their shared a bi-level framework for AAD tasks. Below we present two key distinctions to clarify the unique position of MetaBBO: 
1) Problem Scope and Task Innovation: HH aims to select/generate heuristics within combinatorial optimization (COPs)~\cite{HH-app-1,HH-app-2}, while MetaBBO exclusively targets automating `BBO' algorithm design. Because of its specialized focus, MetaBBO actively samples and models problem landscapes to learn latent optimization dynamics, thereby enabling novel AAD tasks like dynamic algorithm configuration and generation, which remain largely unexplored in HH research~\cite{HH-generation-survey} due to its rigid heuristic-driven framework. 2) Learning Flexibility: Consider the meta-level method, HH primarily relies on meta-heuristics, while MetaBBO integrates diverse ML approaches (reinforcement learning, neuroevolution, supervised learning and in-context learning with LLMs). MetaBBO's neural network-based policies enable \textit{online adaptation} across problem classes, while HH typically operates \textit{offline} within a fixed domain. Specifically, the neural controller of MetaBBO actively interrogates problem characteristics to synthesize algorithm specifically for the current problem landscape, which enables on-instance adjustment as well as cross-problem generalization.

\section{Categorization of MetaBBO by Meta Tasks}\label{sec:3}

\begin{table*}[!ht]
    \centering
    \caption{Representative works in MetaBBO, categorized by different algorithm design tasks. We have provided an \href{https://github.com/MetaEvo/Awesome-MetaBBO}{online page}, where more details of existing MetaBBO works are included. }
    \label{tab:low-level}
    \scalebox{0.80}{
    \begin{threeparttable}
    \begin{tabular}{>{\centering}m{0.8cm}|c|c|>{\centering}m{1.6cm}|>{\centering}m{1.5cm}|m{12cm}}
    \hline
         & \textbf{Algorithm} & \textbf{Year} & \textbf{Low-level Optimizer} & \textbf{Optimization Type} & \textbf{Technical Summary} \\ \hline
        \multirow{11}{*}{\rotatebox{90}{Algorithm Selection}} 
        & Meta-QAP~\cite{as-add-1} & 2008 & MMAS & CO & per-instance algorithm selection by MLP classifier for Quadratic Assignment Problem (QAP) \\ \cline{2-6}
        & Meta-TSP~\cite{as-add-7} & 2011 & GA & CO & per-instance algorithm selection by MLP classifier for Travelling Salesman Problem (TSP) \\ \cline{2-6}
        & Meta-MOP~\cite{as-add-4} & 2019 & MOEA & MOOP & per-instance algorithm selection by SVM classifier from ten multi-objective optimizers \\ \cline{2-6}
        & Meta-VRP~\cite{as-add-5} & 2019 & MOEA & CO & per-instance algorithm selection by MLP classifier from four multi-objective optimizers \\ \cline{2-6}
        & AR-BB~\cite{tianye-bench} & 2020 & EAs, SI & SOP & per-instance algorithm selection by symbolic problem representation and LSTM autoregressive prediction\\ \cline{2-6}
        & ASF-ALLFV~\cite{as-add-3} & 2022 & EAs, SI & SOP & per-instance algorithm selection by adaptive local landscape feature and KNN classifier \\ \cline{2-6}
        & AS-LLM~\cite{2024as-llm} & 2024 & - & SOP & per-instance algorithm selection leverage embedding layer in LLMs \\ \cline{2-6}
        & HHRL-MAR~\cite{as-add-6} & 2024 & SI & SOP & dynamically switch SI optimizers along the optimization process with a Q-table RL agent \\ \cline{2-6}      
        ~ & R2-RLMOEA~\cite{tahernezhad2024r2-rlmoea} & 2024 & EAs & MOOP & dynamically switch 5 EA optimizers along the optimization process with an MLP RL agent \\ \cline{2-6}
        ~ & RL-DAS~\cite{2024rldas} & 2024 & DE & SOP & dynamically switch 3 DE optimizers along the optimization process with an MLP RL agent \\ \cline{2-6}
        ~ & TransOptAS~\cite{cenikj2024transoptas} & 2024 & EAs, SI & SOP & per-instance algorithm selection by Transformer performance predictor from single-objective optimizers \\ \hline

        \multirow{47}{*}{\rotatebox{90}{Algorithm Configuration}} 
        & RLMPSO~\cite{samma2016rlmpso} & 2016 & PSO & SOP & dynamically select PSO update rules \\ \cline{2-6}
        ~ & RL-MOEA/D~\cite{ning2018rlmoea} & 2018 & MOEA/D & MOOP & dynamically control the neighborhood size and the mutation operators used in MOEA/D \\ \cline{2-6}
        ~ & QL-(S)M-OPSO~\cite{liu2019qlmopso} & 2019 & PSO & SOP,MOOP & dynamically control the parameters of PSO update rule \\ \cline{2-6}
        ~ & DE-DDQN~\cite{2019deddqn} & 2019 & DE & SOP & mutation operator selection in DE \\ \cline{2-6}
        ~ & DE-RLFR~\cite{2019derlfr} & 2019 & DE & MMOOP & mutation operator selection in DE for multi-modal multi-objective problems \\ \cline{2-6}
        ~ & LTO~\cite{shala2020lto} & 2020 & CMA-ES & SOP & dynamically configure the mutation step-size in CMA-ES \\ \cline{2-6}
        ~ & QLPSO~\cite{2020qlpso} & 2020 & PSO & SOP & dynamically control the inter-particle communication topology of PSO \\ \cline{2-6}
        ~ & LRMODE~\cite{huang2020lrmode} & 2020 & DE & MOOP & incorporate landscape analysis to operator selection \\ \cline{2-6}
        ~ & RLDE~\cite{hu2021rlde} & 2021 & DE & SOP & dynamically adjust the scaling factor $F$ in DE \\ \cline{2-6}
        ~ & LDE~\cite{sun2021lde} & 2021 & DE & SOP & use LSTM to adaptively control $F$ and $CR$ in DE \\ \cline{2-6}
        ~ & RLEPSO~\cite{yin2021rlepso} & 2021 & PSO & SOP & dynamically adjust factors in EPSO \\ \cline{2-6}
        ~ & qlDE~\cite{huynh2021qlde} & 2021 & DE & SOP & dynamically determine parameter combinations of $F$ and $Cr$ in DE \\ \cline{2-6}
        ~ & DE-DQN~\cite{2021dedqn} & 2021 & DE & SOP & mutation operator selection \\ \cline{2-6}
        ~ & RL-PSO~\cite{wu2022rl-pso} & 2022 & PSO & SOP & dynamically adjust random values in PSO update rule \\ \cline{2-6}
        ~ & RLLPSO~\cite{wang2022rllpso} & 2022 & PSO & LSOP & adaptively adjust the number of performance levels in the population. \\ \cline{2-6}
        ~ & MADAC~\cite{xue2022madac} & 2022 & MOEA/D & MOOP & dynamically adjust all parameters in MOEA/D by an multi-agent system \\ \cline{2-6}
        ~ & RL-CORCO~\cite{hu2022rl-corco} & 2022 & DE & COP & operator selection in constrained problems \\ \cline{2-6}
        ~ & MOEA/D-DQN~\cite{tian2022modead-dqn} & 2022 & MOEA/D & MOOP & leverage DQN to select variation operators in MOEA \\ \cline{2-6}
        ~ & RL-SHADE~\cite{2022rlshade} & 2022 & DE & SOP & perform mutation operator selection in SHADE \\ \cline{2-6}
        ~ & RL-HPSDE~\cite{2022rlhpsde} & 2022 & DE & SOP & control parameter sampling method and mutation operator selection \\ \cline{2-6}
        ~ & NRLPSO~\cite{li2023nrlpso} & 2023 & PSO & SOP & dynamically adjust learning paradigms and acceleration coefficients \\ \cline{2-6}
        ~ & Q-LSHADE~\cite{qlshade} & 2023 & DE & SOP & dynamically control when to use the scheme to reduce the population. \\ \cline{2-6}
        ~ & LADE~\cite{liu2023lade} & 2023 & DE & SOP & leverage three LSTM models to generate three sampling distributions of key parameters in DE \\ \cline{2-6}
        ~ & LES~\cite{2023les} & 2023 & CMA-ES & SOP & use self-attention mechanism to adjust the step size in CMA-ES \\ \cline{2-6}
        ~ & RLAM~\cite{2023rlam} & 2023 & PSO & SOP & enhance the PSO convergence by using RL to control the coefficients of the PSO \\ \cline{2-6}
        ~ & MPSORL~\cite{meng2023mpsorl} & 2023 & PSO & SOP & adaptivly select strategy in multi-strategy PSO \\ \cline{2-6}
        ~ & RLDMDE~\cite{yang2024rldmde} & 2023 & DE & SOP & adaptively select mutation strategy of each population in multi-population DE \\ \cline{2-6}
        ~ & RLMMDE~\cite{han2023rlmmde} & 2023 & MOEA & MOOP & dynamically determine whether to perform reference point adaptation method \\ \cline{2-6}
        ~ & MARLABC~\cite{zhao2023marlabc} & 2023 & ABC & SOP & dynamically select optimization strategy \\ \cline{2-6}
        ~ & CEDE-DRL~\cite{hu2023cede-drl} & 2023 & DE & COP & dynamically select suitable parent population \\ \cline{2-6}
        ~ & AMODE-DRL~\cite{li2023amode-drl} & 2023 & MODE & MOOP & two RL agents, one for mutation operator selection, one for parameter tuning\\ \cline{2-6}
        ~ & RLHDE~\cite{peng2023rlhde} & 2023 & DE & SOP & use Q-learning to select mutation operators in QLSHADE and control the trigger parameters in HLSHADE \\ \cline{2-6}
        ~ & GLEET~\cite{ma2024gleet} & 2024 & PSO,DE & SOP & dynamic hyper-parameters tuning based on exploration-exploitation tradeoff features \\ \cline{2-6}
        ~ & RLMODE~\cite{yu2024rlmode} & 2024 & DE & MOOP & dynamically control the key parameters in DE update rule \\ \cline{2-6}
        ~ & RLNS~\cite{2024PSORLNS} & 2024 & SSA,PSO,EO & MMOP & dynamically adjust the subpopulation size \\ \cline{2-6}
        ~ &  ada-smoDE~\cite{zhang2024gradient} & 2024 & DE & SOP & dynamically control the key parameters in DE update rule \\ \cline{2-6}
        ~ & PG-DE~\cite{zhang2024pg-de} & 2024 & DE & SOP & dynamic operator selection \\ \cline{2-6}
        ~ & SA-DQN-DE~\cite{liao2024sa-dqn-de} & 2024 & DE & MMOP & dynamically select proper local search operators \\ \cline{2-6}
        ~ & RLEMMO~\cite{lian2024rlemmo} & 2024 & DE & MMOP & dynamically select DE mutation operators \\ \cline{2-6}
        ~ & MRL-MOEA~\cite{wang2024mrl-moea} & 2024 & MOEA & MOOP & dynamically select crossover operator in MOEA \\ \cline{2-6}
        ~ & MSoRL~\cite{wang2024msorl} & 2024 & PSO & LSOP & automatically estimate the search potential of each particle \\ \cline{2-6}
        ~ & UES-CMAES-RL~\cite{2024uescmaesrl} & 2024 & UES CMAES & SOP & determine parameters in restart strategy by RL agent \\  \cline{2-6}
        ~ & HF~\cite{pei2024learning} & 2024 & DE & SOP,CO & dynamically select DE mutation operators by RL agent or manual mechanism \\ \cline{2-6}
        ~ & MTDE-L2T~\cite{2024meta-mto} & 2024 & DE & MTOP & control information sharing in multi-population DE to solve MTOP \\ \cline{2-6} 
        ~ & ConfigX~\cite{configx} & 2025 & DE,PSO,GA & SOP & universally control parameters and select operators for modular algorithms \\ \cline{2-6}
        ~ & MetaDE~\cite{metade} & 2025 & DE & SOP & a self-referential framework where DE is used to configure DE parameters \\ \cline{2-6}
        ~ & KLEA~\cite{klea} & 2025 & MOEA & LSMOP & dynamically switch dimension reduction strategies\\  \cline{2-6}
        ~ & RLDE-AFL~\cite{rldeafl} & 2025 & DE & SOP & dynamically select DE mutation and crossover operators for each individual, and control their parameters \\ \cline{2-6}
        ~ & SurrRLDE~\cite{surrrlde} & 2025 & DE & SOP & using Kan-based neural networks as surrogate models for MetaBBO\\ \cline{2-6}
        ~ & LCC-CMAES~\cite{rlcc} & 2025 & CMA-ES & LSOP & dynamically select problem decomposition operators during the cooperative co-evolution  \\\hline
        
        \multirow{17}{*}{\rotatebox{90}{Solution Manipulation}} & RNN-OI~\cite{2017rnnoi} & 2017 & - & SOP & use RNN as a BBO algorithm to output solutions iteratively\\ \cline{2-6}
        ~ & RNN-Opt~\cite{2019rnnopt} & 2019 & - & SOP & using RNN as a algorithm to output sample distribution iteratively \\ \cline{2-6}
        ~ & LTO-POMDP~\cite{2021ltopomdp} & 2021 & - & SOP & LSTM-based optimizer to output per-dimensional distribution \\ \cline{2-6}
        ~ & MELBA~\cite{chaybouti2022melba} & 2022 & - & SOP & use Transformer-based model to output sample distribution \\ \cline{2-6}
        ~ & LGA~\cite{2023lga} & 2023 & GA & SOP & use attention mechanism to imitate crossover and mutation in GA \\ \cline{2-6}
        ~ & OPRO~\cite{yang2024opro} & 2023 & - & SOP & use LLMs as optimizer to output solutions \\ \cline{2-6}
        ~ & LMEA~\cite{liu2024lmea} & 2023 & - & SOP & use LLMs to select parent solutions and perform crossover and mutation to generate offspring solutions \\ \cline{2-6}
        ~ & MOEA/D-LLM~\cite{liu2023moea/d-llm} & 2023 & MOEA/D & MOOP & use LLMs as the optimizer in MOEA/D process \\ \cline{2-6}
        ~ & ELM~\cite{openelm} & 2023 & - & CO & use LLM agent to generate benchmark programs through evolution of existing ones \\ \cline{2-6}
        ~ & ToLLM~\cite{guo2023towards} & 2023 & - & SOP & prompt LLMs to generate solutions \\ \cline{2-6}
        ~ & GLHF~\cite{2024glhf} & 2024 & DE & SOP & use neural network to imitate mutation and crossover in DE\\ \cline{2-6}
        ~ & B2Opt~\cite{li2023b2opt} & 2024 & GA & SOP & use neural network to imitate operators in GA \\ \cline{2-6}
        ~ & RIBBO~\cite{2024ribbo} & 2024 & - & SOP & use GPT model to output optimization trajectories \\ \cline{2-6}
        ~ & EvoLLM~~\cite{2024evollm} & 2024 & - & SOP & imitate ES's optimization behaviour by  iteratively prompting LLM\\ \cline{2-6}
        ~ & EvoTF~\cite{2024evotf} & 2024 & - & SOP & use Transformer-based network to output ES's distribution parameter \\ \cline{2-6}
        ~ & LEO~\cite{brahmachary2024leo} & 2024 & - & SOP & exploitation via LLM instead of crossover and mutation \\ \cline{2-6}
        ~ & CCMO-LLM~\cite{wang2024ccmo-llm} & 2024 & - & CMOP & use LLM as the search operator within a classical CMOEA framework \\ \hline

        \multirow{10}{*}{\rotatebox{90}{Algorithm Generation}} 
        & GSF~\cite{yi2022GSF} & 2022 & - & CO & generate whole BBO algorithm by using RL agent to select operators from fixed algorithmic template \\ \cline{2-6}
        ~ & AEL~\cite{liu2023ael} & 2023 & - & CO & use LLM to evolve algorithm source code \\ \cline{2-6}
        ~ & EoH~\cite{liu2024eoh} & 2023 & - & CO & use LLM agent to evolve algorithm's thoughts and source code \\ \cline{2-6}
        ~ & SYMBOL~\cite{2024symbol} & 2024 & - & SOP & automatically generate symbolic update rules along optimization process through LSTM \\ \cline{2-6}
        ~ & LLaMEA~\cite{2024llamea} & 2024 & - & SOP & use LLM to evolve EA algorithm \\ \cline{2-6}
        ~ & LLMOPT~\cite{huang2024llmopt} & 2024 & - & MOOP & use LLM to evolve operators for multi-objective optimizer \\ \cline{2-6}
        ~ & LLaMoCo~\cite{2024llamoco} & 2024 & - & SOP & instruction-tuning for LLM to generate accurate algorithm code \\ \cline{2-6}
        ~ & OptiMUS~\cite{ahmaditeshnizi2023optimus} & 2024 & - & MILP & develop multi-agent pipelines for LLM to solve MILP problem as a professional team \\ \cline{2-6}
        ~ & LLM-EPS~\cite{zhang2024understanding} & 2024 & - & - & use LLM to generate offspring codes in evolutionary program search \\ \cline{2-6}
        ~ & ALDes~\cite{zhao2024aldes} & 2024 & - & SOP & sequentially generate each component in an algorithm through auto-regressive inference\\ \hline
        
    \end{tabular}
    \end{threeparttable}}
\end{table*}

We introduce four common meta-level tasks in MetaBBO: Algorithm Selection in Section~\ref{sec:as}, Algorithm Configuration in Section~\ref{sec:ac},  Solution Manipulation in Section~\ref{sec:ai}  and Algorithm Generation in Section~\ref{sec:ag}. Generally speaking, they are organized by the size of design space, from smallest to largest. 
Algorithm selection deals with a small space, choosing from a few BBO optimizers, while algorithm generation explores a vast space, allowing the meta-level policy to create novel BBO optimizers in an open-ended manner. Table~\ref{tab:low-level} presents a selection of works categorized by the meta tasks, along with their references, publication years, low-level optimizers, targeted problem types\footnote{We use SOP, MOOP, COP, CMOP, MMOP, MMOOP, LSOP, LS-MOOP, MILP and CO to denote single-objective optimization, multi-objective optimization, constrained optimization, constrained multi-objective optimization, multi-modal optimization, multi-modal multi-objective optimization, large-scale optimization, large-scale multi-objective optimization, mixed integer linear programming and combinatorial optimization, respectively.}, and technical summaries. 

\subsection{Algorithm Selection}\label{sec:as}
\begin{figure}[t]
    \centering
    \includegraphics[width=0.8\linewidth]{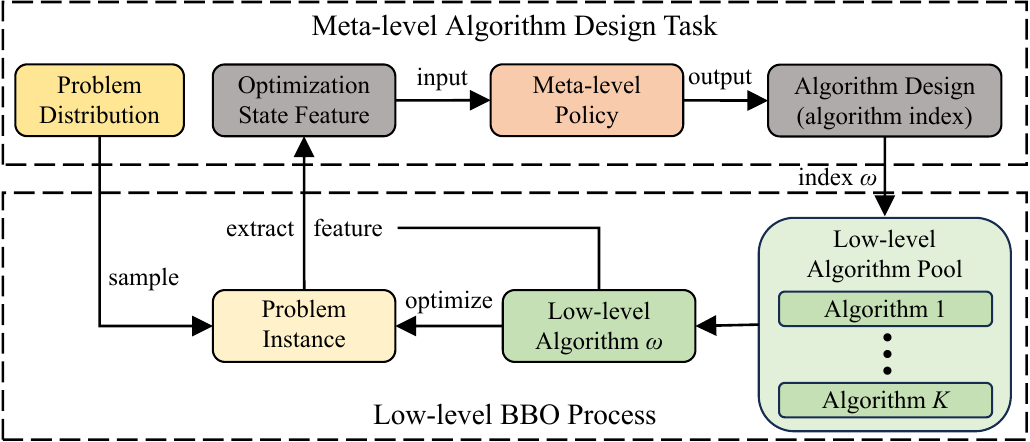}
    \caption{Conceptual workflow of MetaBBO for Algorithm Selection.}
    \label{fig:metabbo_aas}
\end{figure}

Algorithm Selection (AS) has been discussed for decades~\cite{2019aas-survey,cenikj2024survey}. The goal of AS is to select the most suitable algorithm from the algorithm pool according to the target task. The motivation of AS is that optimization behaviors and preferred scenarios vary with the algorithms, resulting in a notable performance difference~\cite{1976aas-survey2}. Initially, AS is performed by human experts, who suggest algorithms based on their knowledge, which is labor-intensive and requires extensive expertise. To alleviate this dependency, researchers seek to develop more automated approaches. 

\subsubsection{\textbf{Formulation}}
We examine the common AS paradigm. In the low-level BBO process, the component $\mathcal{A} = \{\mathcal{A}_1,...\mathcal{A_\text{K}}\}$ represents an algorithm pool $\mathcal{A}$ with $K$ candidate BBO algorithms. The algorithm design space $\Omega=\{1,2...,K\}$ is the selective space involving all indexes of the candidate algorithms, where $\omega \in \Omega$ denotes an index of a candidate from $\mathcal{A}$. For each problem instance $f_i$ in the training set, the goal of AS is to output an algorithm decision $\omega_i^t$ for $f_i$ at each optimization step $t$. As illustrated in Fig.~\ref{fig:metabbo_aas}, MetaBBO automates this task by maintaining a learnable meta-level policy $\pi_\theta$ with parameters $\theta$, which takes a state feature $s_i^t$ obtained by $\text{sf}(\cdot)$ describing the optimization state of this optimization step, and then outputs $\omega_i^t$. The selected candidate algorithm $\mathcal{A}[\omega_i^t]$ is used to optimize $f_i$ in the low-level BBO process. Its performance on $f_i$ serves as the performance measurement in Eq.~(\ref{eq:1}). MetaBBO aims to find an optimal meta-level policy that suggests a best-performing algorithm in $\mathcal{A}$ for each $f_i$ at each optimization step $t$ automatically. 
Suppose the optimization horizon of the low-level BBO process is $T$, the meta-objective $J(\theta)$ of AS is calculated as:
\begin{equation}\label{eq: as2}
    J(\theta) \approx \frac{1}{N}\sum_{i=1}^{N}\sum_{t=1}^T \text{perf}(\mathcal{A}[\omega_i^t], f_i)
\end{equation}
After training, $\pi_\theta$ is expected to select well-matched candidate algorithms from $\mathcal{A}$ for unseen problems.     
\subsubsection{\textbf{Related Works}}
First, per-instance AS is widely adopted in the literature, where a single algorithm is selected for the entire optimization progress for each specific problem, meaning that $\omega_i^t$ remains time-invariant. 
A straightforward approach to learning an effective meta-level policy for the AS task is to form a logical association between the attributes of $f_i$ and the algorithm selection decision $\omega_i$ that corresponds to them. Since typically the number of candidate algorithms in the pool $\mathcal{A}$ is finite, many early-stage MetaBBO for AS researches transformed the meta-level learning process to a classification task~\cite{bischl2012init-as-ela, 2019as-combine-ela-ml,as-add-1,as-add-7,as-add-4,as-add-5,as-add-3,tianye-bench,2024as-llm}. In their methodologies, the state feature extraction function $\text{sf}(\cdot)$ in Eq.~(\ref{eq:1}) extracts problem characteristics $s_i$ of $f_i$, which is significant enough to distinguish $f_i$ with the other problem instances. A benchmarking process is employed to identify the top-performing candidate algorithm for $f_i$. The identified algorithm is then used as the classification label. The meta-level policy $\pi_\theta$ is regarded as a classifier and hence meta-trained to achieve maximum prediction accuracy. The state feature extraction mechanism $\text{sf}(\cdot)$ in these works can be very different according to the target optimization problem types. Meta-QAP~\cite{as-add-1}, Meta-TSP~\cite{as-add-7} and Meta-VRP~\cite{as-add-5} construct an information collection termed as meta data for combinatorial optimization problems, which maintains the nodes information, edge connections in the graph and constraints of a problem instance. For continuous single-objective optimization problems, exploratory landscape analysis (ELA) techniques are adopted in~\cite{bischl2012init-as-ela,2019as-combine-ela-ml,as-add-4,as-add-3}, which profiles the objective space characteristics of a problem instance such as the convexity, peaks and valleys. While for multi-objective optimization, the options are comparatively limited, with representative works as the decomposition-based landscape features~\cite{moo-landscape-sv1, moo-landscape-1}. These works mainly apply basic classification models such as Support Vector Machine (SVM), K-Nearest Neighbors (KNN) and Multi-Layer Perceptron (MLP) for the label prediction. In contrast, to achieve in-depth data mining of the relationship between the problem structures and the optimizer performance, the study in~\cite{tianye-bench} uses symbolic regression techniques to recover the mathematical equation of the given problem and then leverages a Long Short-Term Memory (LSTM)~\cite{hochreiter1997lstm} to auto-regressively predict the desired candidate algorithm. 
The study in \cite{Quentin2024utility} introduces time-series of fitnesses obtained from the first few iterations of an algorithm as trajectory feature inputs and employs ML methods such as Random Forest for algorithm selection. Ana et al.~\cite{perrun,jankovic2022trajectory} propose per-run AS that conducts a warm-starting strategy to enhance the trajectory-based ELA sampling.  
TransOptAS~\cite{cenikj2024transoptas} explores the possibility of constructing a performance indicator based solely on the raw objective values to eliminate the computation cost for computing sf$(\cdot)$. It leverages a Transformer~\cite{transformer}-styled architecture that takes a batch of sampled objective values as input and outputs the performance of the candidate algorithms through supervision under the benchmark results. 
AS-LLM~\cite{2024as-llm} leverages pre-trained LLM embeddings to extract features from the candidate algorithms and the target optimization problem, then selects the best algorithm by feature similarities.
Several latest MetaBBO works explored the possibility of extending per-instance AS to dynamic AS during the low-level BBO process~\cite{as-add-6,tahernezhad2024r2-rlmoea,2024rldas}. Concretely, the meta-level algorithm design task in this paradigm turns to flexibly suggest one candidate algorithm to optimize $f_i$ for each optimization step $t$. 
The dynamic AS is regarded as Markov Decision Process in the mentioned MetaBBO works and hence can be maximized by using RL to meta-learn an optimal policy. The optimization state feature in RL-DAS~\cite{2024rldas} includes not only the problem properties but also the dynamic optimization state information to support such flexible algorithm switch which help RL-DAS achieve at most 13\% performance improvement over the advanced DE variants in its algorithm pool.

Moreover, the fundamental nature of AS highlights the significance of constructing the algorithm pool $\mathcal{A}$, which necessitates a comprehensive understanding of the target problem distribution and effective BBO algorithms. An effective pool should include a variety of BBO algorithms to tackle problems with diverse characteristics. Consequently, there is considerable interest among researchers in the automatic construction of algorithm portfolios. This involves leveraging data-driven approaches, such as Hydra~\cite{hydra}, AutoFolio~\cite{autofolio}, and PS-AAS~\cite{PS-AAS}, which offer extensive analysis and valuable insights for developing the algorithm pool $\mathcal{A}$.

\subsubsection{\textbf{Challenges}}
While past research has made progress in AS, several technical challenges persist:
\begin{itemize}
    
    \item 
    For per-instance AS, labeling the training set is expensive due to the exhaustive search needed to find the optimal algorithm for each instance. Limited candidates and problem instances lead to generalization issues. In dynamic AS, the increased methodological complexity challenges the learning effectiveness of RL methods.
    \item 
    The algorithm design space in MetaBBO for AS is coarse-grained, limited by the performance of individual algorithms without tuning their configurations. In the next section, we introduce algorithm configuration tasks, which offer larger and more fine-grained design spaces.
\end{itemize}   

\subsection{Algorithm Configuration}\label{sec:ac}
Algorithm configuration~(AC) is a key task in optimization, since almost all BBO algorithms possess hyper-parameters \cite{2021hpob} and optional operators~\cite{2010os-survey} that affect performance. 
To automate the AC task, various adaptive and self-adaptive BBO algorithms have been developed in the past decades~\cite{2022dac-survey}. Algorithms like JADE~\cite{2009jade} and APSO~\cite{2009apso} leverage historical optimization data to compute informative decision statistics such as the potential of the hyper-parameter values and the success rates of the optional operators~\cite{2021madde,2021jde,2020imode}. 
However, as discussed in the introduction, these approaches suffer from design bias, limited generalization, and high labor costs.
\subsubsection{\textbf{Formulation}}
\begin{figure}[t]
    \centering
    \includegraphics[width=0.8\linewidth]{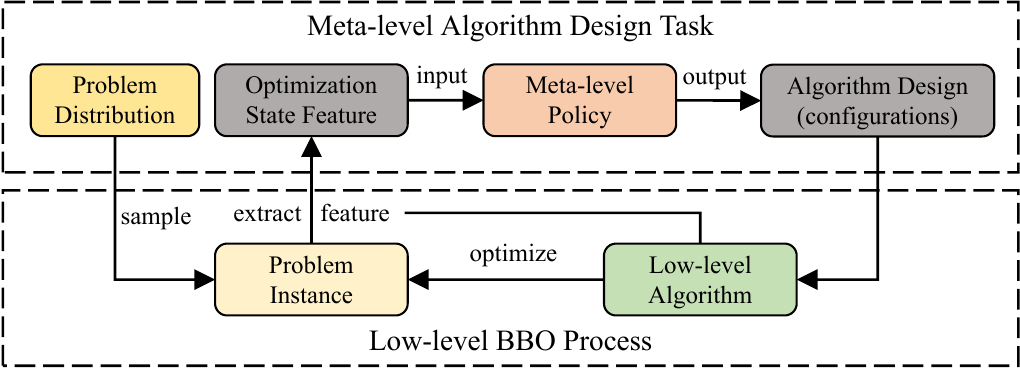}
    \caption{Conceptual workflow of MetaBBO for Algorithm Configuration.}
    \label{fig:metabbo_dac}
\end{figure}

As shown in Fig.~\ref{fig:metabbo_dac}, MetaBBO overcomes the limitations of manual AC techniques by using meta-learning to develop a meta-level configuration policy. This policy dynamically adjusts a BBO algorithm throughout the lower-level BBO procedure. More formally: in the low-level BBO process, the optimizer $\mathcal{A}$ represents the BBO algorithm to be configured. The algorithm design space $\Omega$ is hence the configuration space of $\mathcal{A}$. The size of $\Omega$ can be either infinite~(with continuous hyper-parameters) or finite~(with discrete hyper-parameters or several optional operators). MetaBBO dictates AC in a dynamic manner: given a problem instance $f_i$, at each optimization step $t$ of the low-level BBO process, a state feature $s_i^t$ is obtained by $\text{sf}(\cdot)$ to describe the state of this optimization step. 
The meta-level policy $\pi_\theta\left(s_i^t\right)$ outputs the algorithm design $\omega_i^t$, which sets the configuration of $\mathcal{A}$ as $\mathcal{A}$.set $\left(\omega_i^t\right)$. Then the algorithm is used to optimize $f_i$ for the current optimization step. 
Suppose the optimization horizon of the low-level BBO process is $T$, the meta-objective $J(\theta)$ of MetaBBO for AC is formulated as

\begin{equation}\label{eq: ac}
    J(\theta) \approx \frac{1}{N}\sum_{i=1}^{N}\sum_{t=1}^T \text{perf}(\mathcal{A}.set(\omega_i^t), f_i)
\end{equation}

MetaBBO for AC improves on human-crafted adaptive methods by meta-learning the configuration policy through optimizing the meta-objective in Eq.~(\ref{eq: ac}), removing the need for labor-intensive, expert-driven designs. The bi-level meta-learning paradigm also enhances generalization, as the policy can be trained on a large set of problem instances, distilling configuration strategies that can be applied to new problems.

\subsubsection{\textbf{Related Works}}
Typically, the AC studies involve a two-step process: initially choosing an algorithm template and subsequently adjusting internal components or parameters. In this paper, we categorize existing MetaBBO for AC research into three distinct subgroups based on the second step. The first sub-category is adaptive  operator selection~(AOS), where several optional operators is flexibly selected by the meta-level policy. The second is hyper-parameter optimization~(HPO), where the hyper-parameter values are controlled by the meta-level policy. The last is the combination of AOS and HPO, where $\Omega$ is a complex configuration space including both hyper-parameters and operators. Next we introduce related works in these sub-categories.  

\paragraph{Adaptive Operator Selection}
The works in this line aims to dynamically switch the operators of the low-level BBO algorithms during the optimization process. 
The majority of AOS methods still focus on DE algorithms~\cite{2021dedqn,sallam2020marlwcma,huang2020lrmode,2019deddqn,2022rlshade,hu2022rl-corco,tian2022modead-dqn,peng2023rlhde,yang2024rldmde}, due to their strong performance and the availability of various operators for selection. 
These works share similar methodologies: a mutation operator pool is maintained, involving representative mutation operators such as DE/rand/2, DE/best/2, DE/current-to-rand/1, DE/current-to-best/1 and DE/current-to-pbest/1. In order to address different types of problems, the technical differences in these works revolve around the tailored state feature extraction design and the operator pool. The state feature extraction in these works can be divided into two main strategies: discrete representation and continuous representation. 

For discrete state representation, the study in~\cite{sallam2020marlwcma} first computes the diversity variation and the performance improvement between two consecutive optimization steps as an effective profile of the optimization dynamics. These two indicators, being continuous variables, are then divided into five distinct levels each. According to the discretized state feature, a Q-table policy is constructed to select one operator from an operator pool with three candidates.  RLHDE~\cite{peng2023rlhde} uses the relative density in the solution space and the objective space against the initial population and objective values to indicate the convergence trend and the performance improvement. The values of the two density indicators are discretized into five and four levels respectively, constituting $20$ different optimization states. 
The operators pool in RLHDE involves six mutation operators, which improve the diversity of the optimization behaviours. 
RL-CORCO~\cite{hu2022rl-corco} addresses constrained multi-objective optimization by enhancing the CORCO algorithm through multiple Q-table policies. In the algorithm, each sub-population maintains a Q-table, where rows represent nine states indicating different levels of objective improvement and constraint violation, and columns represent two mutation operators. The policy selects the appropriate mutation operator to optimize the solution as effectively as possible.

Compared to discrete features, continuous state feature extraction enables finer state modeling, providing unique representations for optimization states and leading to smarter decisions by the meta-level policy.
For instance, DE-DDQN~\cite{2019deddqn} proposes a very comprehensive optimization state extraction function, which computes a total of $99$ features: the first $19$ features describe the optimization progress and the properties of the target optimization problems, while the rest $80$ are statistics describing the optimization potential of the four mutation operators in the operator pool. An MLP neural network-based meta-level policy generates Q-values for the candidate mutation operators and the one with maximal Q-value is chosen for the next optimization step. Following DE-DDQN, DEDQN~\cite{2021dedqn} and MOEA/D-DQN~\cite{tian2022modead-dqn} also construct MLP policies. DEDQN indicates that the features in DE-DDQN show certain redundancy and might fall short in capturing the local landscape features. To address this, DEDQN proposes a feature extraction mechanism inspired from classical fitness landscape analysis~\cite{fla}. 
By using random walk sampling, DEDQN computes ruggedness and fitness distance correlations in the local landscape. 
Results show that landscape features are effective for MetaBBO methods to generalize across problem types, i.e., from synthetic problems to realistic problems~\cite{2024metabox}.
For addressing multi-objective optimization problem, MOEA/D-DQN embeds the information of the reference vectors in MOEA/D into the state extraction. KLEA~\cite{klea} further explore effective RL policy that could adaptively select desired dimension reduction strategies to enhance MOEA/D in large scale problem instances. 
To tackle multi-modal optimization problem, RLEMMO~\cite{lian2024rlemmo} first clusters solutions to compute the neighborhood information. The optimization state is then constructed by concatenating optimization progress, distributional properties and neighborhood information.

Besides the selection of DE modules~\cite{modde}, other BBO algorithms such as PSO and CMA-ES also have their own modular frameworks (i.e., PSO-X~\cite{psox} for PSO and modCMA~\cite{modCMA} for CMA-ES), and operator selection methods~\cite{samma2016rlmpso,meng2023mpsorl}. Furthermore, a novel genetic BBO algorithm modularization system covering diverse algorithm modules from DE, PSO and GA is developed in~\cite{configx}, where a Transformer-based agent, named ConfigX, is proposed to meta-learn a universal configuration policy through multitask reinforcement learn-
ing across a designed joint optimization task space. 

\paragraph{Hyper-parameter Optimization}
Several early attempts meta-learn a configuration policy that dictates a single hyper-parameter setting throughout the entire process of solving a problem instance~\cite{2023rl-piac,2011lkh}. Now, most MetaBBO for AC approaches follow the dynamic AC paradigm in Eq.~(\ref{eq: ac}), offering a flexible exploration-exploitation tradeoff to further improve the optimization performance. Since different BBO algorithms have distinct hyper-parameters, existing MetaBBO for AC works customize their methods to explore the intricate relationships between the hyper-parameters  and the resulting exploration-exploitation tradeoff in each specific algorithm.

Since DE is known to be highly sensitive to hyperparameter settings, particularly the scaling factor 
$F$ and the crossover probability $Cr$, many efforts have focused on meta-tuning DE. RLDE~\cite{hu2021rlde} propose a simple Q-table policy to adjust $F$ when optimizing the power generation efficiency in solar energy system. It uses a Boolean indicator as the optimization state feature: indicating whether the solution quality is improved between two optimization steps. The algorithm design space, represented as $\delta F \in \{-0.1,0,0.1\}$, indicates the variation in $F$ for the subsequent optimization step. Following RLDE, QLDE~\cite{huynh2021qlde} extends the algorithm design space to five combinations of the parameter values. RLMODE~\cite{yu2024rlmode} further proposes a specific state extraction function for constrained multi-objective optimization, which divided the state feature into eight possible situations, according to the solution feasibility and the dominance relationship. The algorithm design space is three combinations of different $F$ and $Cr$ settings to represent different exploration-exploitation tradeoffs. The Q-table agent is updated by first selecting the most promising combination and then observing the resulting performance improvement. For more fine-grained parameter control, LDE~\cite{sun2021lde} firstly considers using recurrent neural network~(i.e., LSTM) as the meta-level policy, which extracts hidden state feature for separate optimization step and outputs the values for $F$ and $Cr$ from a continuous range $[0,1]$. 
The same authors subsequently propose LADE~\cite{liu2023lade} as a extension of LDE. Compared to LDE, LADE aims to control more hyper-parameters including not only the mutation strength and crossover rate but also the update weights. All parameters are represented as matrix operations. Instead of using one LSTM for controlling all parameters, LADE's meta-level policy comprises three LSTM networks for controlling these parameters respectively. LADE show more robust learning effectiveness than LDE.
A recent study, L2T~\cite{2024meta-mto}, employs the MetaBBO framework to regulate the setting of DE parameters and the likelihood of knowledge transfer within the multitask optimization working scenarios. The state feature is represented by the rate of successful transfers and the enhancement in sub-population performance.
Despite adapting $F$ and $Cr$, the control of population size is considered in 
Q-LSHADE~\cite{qlshade}.  The algorithm design space is the decay rate of the linear population size reduction in LSHADE~\cite{lshade}, which can take values from $\{0, 0.2\}$. 
Q-LSHADE also meta-learns a Q-table policy by the feedback indicating the performance improvement. There are also several MetaBBO works which facilitate hyper-parameter optimization on other algorithms, such as PSO~\cite{2020qlpso,wang2022rllpso,2024PSORLNS,liu2019qlmopso,yin2021rlepso,wu2022rl-pso,2023rlam,ma2024gleet}, ES~\cite{shala2020lto,2023les}, and the Firefly algorithm~\cite{2018QFA}. In addition, a recent work GLEET~\cite{ma2024gleet} proposes a general learning paradigm which show generic HPO ability for both DE and PSO. Due to the space limitation, other related works are summarized in Table~\ref{tab:low-level}.

\paragraph{Hybrid Control}Some MetaBBO works explore other AC perspectives~\cite{zhang2024pg-de,pei2024learning}. In particular, the combination of HPO and AOS has gained significant attention~\cite{li2023amode-drl,ning2018rlmoea,peng2023rlhde,2022rlhpsde,li2023nrlpso,xue2022madac}, since learning a meta-level policy in $\Omega_{\text{HPO+AOS}}$ would probably result in a better AC policy than learning them separately. Nevertheless, this poses a significant challenge as learning from an expanded algorithm design space necessitates more intricate learning strategies and model frameworks. Thoughtful design is essential to guarantee effective learning.


\subsubsection{\textbf{Challenges}}
Despite their success, existing MetaBBO works for AC still face some challenges.

\begin{itemize}
    \item A certain proportion of existing methods use a very limited set of training problems. In particular, some only train their meta-level policies on a specific optimization problem instance, raising doubts about the true generalization performance of the resulting policies. 
    \item MetaBBO for AC works operate on the basis of predefined low-level BBO algorithms. Hence, the performance of these methods is closely tied to the original BBO algorithm. Furthermore, the inherent algorithm structures, optimization logic, and design biases significantly restrict the algorithm design space. Can we further expand the algorithm design space and step out this boundary? In the next two subsections, we introduce two novel categories of MetaBBO works that offer potential solutions.  
\end{itemize}

\subsection{Solution Manipulation}\label{sec:ai}
\begin{figure}[t]
    \centering
    \includegraphics[width=0.8\linewidth]{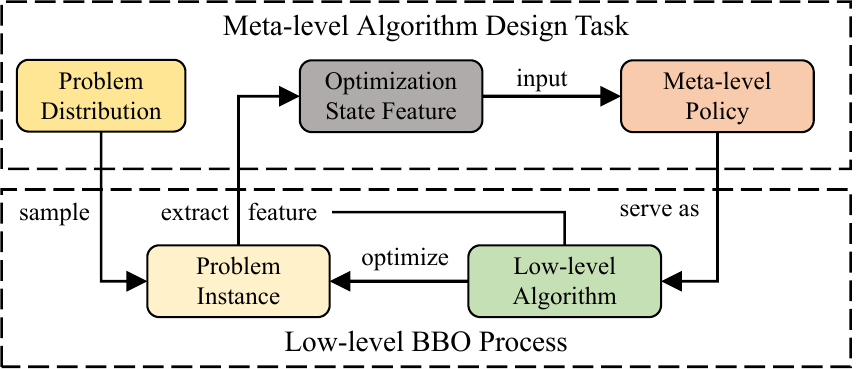}
    \caption{Conceptual workflow of MetaBBO for Solution Manipulation.}
    \label{fig:metabbo_asalg}
\end{figure}

So far, we have introduced two basic categories of MetaBBO: AS and AC. 
An intuitive observation is that within the MetaBBO framework for AS/AC tasks, the low-level BBO procedure necessitates a BBO algorithm as the foundational optimizer, which comes with a defined algorithm design space (e.g., algorithm pool or configuration space). 
This leads to two limitations. First, it requires expert knowledge to select an appropriate BBO algorithm, otherwise the meta-level policy’s learning effectiveness and overall performance may suffer. Second, managing both the meta-level policy and the low-level BBO optimizer simultaneously incurs certain computational costs. To address these limitations, several MetaBBO works have explored the potential of directly using the meta-level policy for solution manipulation. This approach integrates meta-level training and low-level optimization into a single entity, eliminating the need for a predefined BBO algorithm. In this framework, the meta-level policy itself functions as an optimization algorithm, directly manipulating candidate solutions throughout the optimization process. We illustrate this MetaBBO workflow in Fig.~\ref{fig:metabbo_asalg}, referring to it as MetaBBO for solution manipulation~(SM).   

\subsubsection{\textbf{Formulation}}
To formulate the process of solution manipulation in MetaBBO, some clarifications have to be made. First, MetaBBO for SM integrates the functions of meta-level policy and the low-level BBO algorithm into a single parameterized agent $\pi_\theta$, removing the need for a traditionally perceived BBO algorithm. Therefore, the meta-level policy $\pi_\theta$, typically a neural network, inherently serves as the BBO algorithm. In this case, the algorithm design space $\Omega$ turns to the parameter space of the policy, where each algorithm design $\omega$ in this space corresponds to the values of the neural network parameters $\theta$. Given a problem instance $f_i$, at each optimization step $t$, the optimization state feature $s_i^t$ is first computed by $\text{sf}(\cdot)$. According to $s_i^t$, the policy~(acts as the BBO algorithm) $\pi_\theta$ optimizes $f_i$ for one optimization step, e.g., reproducing the candidate solutions. The performance improvement is hence measured as $\text{perf}(\pi_\theta(s_i^t),f_i)$. Suppose the optimization horizon of the low-level BBO process is $T$, the meta-objective of MetaBBO for SM is formulated as
\begin{equation}\label{eq: ai}
    J(\theta) \approx \frac{1}{N}\sum_{i=1}^{N} \sum_{t=1}^T \text{perf}(\pi_\theta(s_i^t), f_i)
\end{equation}
Through maximizing $J(\theta)$ over $N$ problem instances in the training set, a neural network-based BBO algorithm is obtained, functioning similarly to human-crafted BBO algorithms: iteratively optimizes the problem instances. Next, we next introduce representative MetaBBO works for SM.


\subsubsection{\textbf{Related Works}}
An intuitive way of resembling the iterative optimization behaviour by neural networks is considering temporal network structure such as recurrent neural networks~\cite{2017rnnoi,2019rnnopt,2021ltopomdp}, which enable MetaBBO to directly adjust candidate solutions over sequential steps. The corresponding mathematical formulation is quite straightforward:
\begin{equation}~\label{eq:rnnoi}
    X^t, h^t = \pi_\theta(X^{t-1},Y^{t-1},h^{t-1}), \quad Y^t = f_i(X^t) 
\end{equation}
where $\pi_\theta$ is an RNN/LSTM, $h^t$ is the hidden state. This paradigm is first adopted in RNN-OI~\cite{2017rnnoi}, which meta-learns an LSTM to reproduce candidate solutions. For each $f_i$ in the training problem set, RNN-OI randomly initializes a solution $X^0$, obtains the corresponding objective values $Y^0$, and then optimizes $f_i$ by iteratively inferring the next-step solution. To meta-learn a well-performing $\pi_\theta$, the observed improvement per step is computed as the $\text{perf}(\cdot)$ function. 
Once trained, the LSTM serves as a BBO algorithm and iteratively optimizes the target optimization problem following Eq.~(\ref{eq:rnnoi}). 
Due to the end-to-end inferring process, RNN-OI is shown to run $10^4$ times faster compared to hand-crafted algorithms such as Spearmint~\cite{Spearmint}.
Following RNN-OI, similar works include RNN-Opt~\cite{2019rnnopt} improving RNN-OI through input normalization and constraint-dependent loss function, LTO-POMDP~\cite{2021ltopomdp} using neuroevolution to learn the network parameters, MELBA~\cite{chaybouti2022melba} improving the long sequence modelling of RNN/LSTM by introducing Transformer structure, and RIBBO~\cite{2024ribbo} leveraging efficient and generic behaviour cloning framework to learn an optimizer that resembles the given teacher optimizer. 

Nevertheless, the above works still suffer from generalization limitation and interpretability issues. On the one hand, the optimization state features only include the raw population information, which makes the policy easily overfits to the training problems. On the other hand, the learned policies in these works shift toward ``black-box'' systems, which hinders further analysis on what they have learned. In the last two years, several more interpretable MetaBBO for SM works are proposed to address these issues~\cite{li2023b2opt,2023lga,2024evotf,2024glhf}. These works propose using higher-level features as a substitute for the raw features to achieve generalizable state features across diverse problems. Typically, these features include the distributional characteristics of the solution space and the objective space, the rank of objective values, and the temporal features reflecting the optimization dynamics. 
They have proposed several novel architecture designs to make the meta-level policy explicitly resembles representative EC algorithms such GA~\cite{li2023b2opt,2023lga}, DE~\cite{2024glhf}, and ES~\cite{2024evotf}. For instance, LGA~\cite{2023lga} designs two attention-based neural network modules to act as the selection and mutation rate adaption mechanisms in GA. The parameterized selection module applies cross-attention between the parent population and the child population, and the obtained attention score matrix is used as the selection probability.  The parameterized mutation rate adaption module applies self-attention within the child population, and the obtained attention scores is used as the mutation rate variation strength. B2Opt~\cite{li2023b2opt} improves LGA by proposing a novel, fully end-to-end network architecture which resembles all algorithmic components in GA, including crossover, mutation, selection. For example, the selection module within B2Opt utilizes a method similar to the residual connection in Transformer, facilitating the use of matrix operations for selecting populations. 
By meta-training the proposed meta-level policies on the training problem set, these MetaBBO for SM works show competitive optimization performance. In particular, their meta-level policies are trained with low dimensional synthetic problems~($\leq $ 10) yet could be directly generalized for solving high dimensional continuous control problems~($>$ 500)~\cite{2023lga}, e.g., neuroevolution~\cite{ne-bench}.

With the emergence of LLMs, their ability to understand the reasoning in natural language outlines a novel opportunity for SM. Related works in this line widely leverage the In-Context Learning~(ICL)~\cite{ICL} to prompt with general LLMs iteratively as an analog to BBO algorithms to reproduce solutions. A pioneer work is OPRO~\cite{yang2024opro}, which first provides LLMs a context of the problem formulation and historical optimization trajectory described in natural language. It then prompts LLMs to suggest better solutions based on the provided context. This idea soon becomes popular and spreads to multiple optimization scenarios such as program search~\cite{openelm} combinatorial optimization~\cite{liu2024lmea}, multi-objective optimization~\cite{liu2023moea/d-llm,wang2024ccmo-llm}, large scale optimization problem~\cite{2024evollm,brahmachary2024leo} and prompt optimization~\cite{2023evoprompt}. The eye-catching advantage of LLM-based SM is that it requires minimal expertise - users only need to describe the optimization problem in nature language, and LLMs handle the rest.

\subsubsection{\textbf{Challenges}}
As a novel direction, MetaBBO for SM is promising due to the end-to-end manner. However, several technical challenges remain:
\begin{itemize}
    \item Approaches like RNN-Opt directly learn to manipulate candidate solutions without following a specific algorithm structure. While this provides flexibility, these methods often lack transparency and clear understanding of their inner workings. Additionally, due to the complexity of BBO tasks, exploring strong neural networks capable of handling diverse, complex problems remains a challenge.

    \item In contrast, methods like LGA closely mimic the structure and components of existing EAs, making the process more transparent. However, because these methods resemble existing algorithms, their performance might be inherently constrained by the limits of the original ones.

    \item MetaBBO approaches that use LLMs, while reducing the need for manual algorithm design, face significant computational overhead. The iterative interactions with LLMs generate large volumes of tokens, leading to inefficiencies in both time and cost.


    \item Finally, MetaBBO for SM treats the policy itself as the optimizer, targeting at learning the optimal mapping from current landscape to next candidate positions. However, this remains a highly challenging task for continuous BBO tasks. The possible landscapes  are diverse and infinite.  As a result, so far, it is very challenging to build and train a model that can effectively handle these complexities in practice. In the next section, we will explore the ``algorithm generation'' approach, which leverages the meta-level policy as an algorithm discoverer, namely, using learning to create new algorithmic workflows, update rules, and implementations.

\end{itemize}

\subsection{Algorithm Generation}\label{sec:ag}
\begin{figure}[t]
    \centering
    \includegraphics[width=0.7\linewidth]{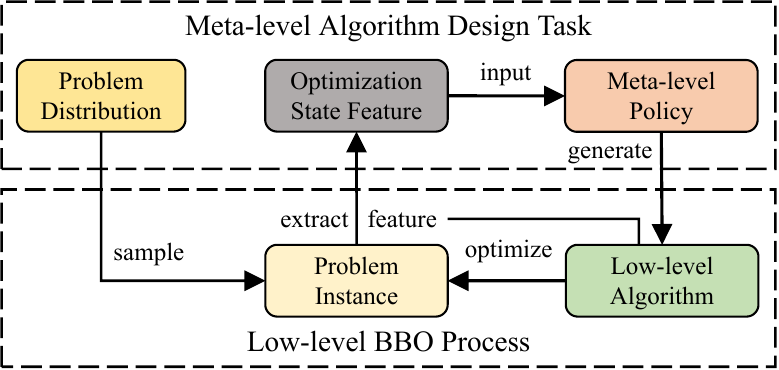}
    \caption{Conceptual workflow of MetaBBO for Algorithm Generation.}
    \label{fig:metabbo_discover}
    \vspace{-3mm}
\end{figure}

Besides MetaBBO for SM, an interesting research question comes out: whether learning-based systems such as MetaBBO could automatically create~(generate) new BBO algorithms with competitive optimization performance and minimal expertise requirement? To this end, MetaBBO for algorithm generation~(AG) presents a different methodology: meta learning a parameterized policy that could discover novel algorithms accordingly without the human-expert prior, of which the workflow is illustrated in Fig.~\ref{fig:metabbo_discover}. The difference between AG and SM is that the meta-level policy in SM plays both the role of the meta-level policy and the low-level optimizer, while the meta-level policy in AG is trained to output a complete optimizer which is used then in the low-level BBO process. 

\subsubsection{\textbf{Formulation}}
MetaBBO for AG works construct an algorithm representation space $\Omega$ as its design space. For example, $\Omega$ can be a algorithm workflow space, a mathematical expression space or a programming language space, reflecting the way humans express algorithms - through modular algorithm workflows, symbolic mathematical expressions or programming language syntax. For a problem instance $f_i$, a concrete algorithm design $\omega_i^t$ is output by the meta-level policy $\pi_\theta$, according to the optimization state feature $s_i^t$. The $\text{sf}(\cdot)$ function, in this case, can incorporate landscape features, symbolic representations, or natural language descriptions of $f_i$. The generated $\omega_i^t$ can be a complete workflow, a mathematical expression or a functional program that represents a novel BBO algorithm $\mathcal{A}$. the meta-objective of MetaBBO for AG is to meta learn a policy $\pi_\theta$ capable of generating well-performing algorithms:
\begin{equation}\label{eq: ag}
    J(\theta) \approx \frac{1}{N}\sum_{i=1}^{N} \sum_{t=1}^{T} \text{perf}(\omega_i^t, f_i), \quad \omega_i^t = \pi_\theta(s_i^t)
\end{equation}
where $\text{perf}(\omega_i^t, f_i)$ is the one-step optimization performance gain of the generated algorithm on $f_i$. Through training the policy across a problem set, the policy is expected to automatically generate flexible and even novel BBO algorithms to address various optimization problems. Besides, note that MetaBBO for AG could work with varying granularity: a)~generating a universal algorithm for all problems~\cite{liu2024eoh}, b)~generating customized algorithms for each problem~\cite{zhao2024aldes}, and c)~generating flexible optimization rules that adapt to each step of the optimization process and each specific problem~\cite{2024symbol,yi2022GSF}. In Eq.~(\ref{eq: ag}), we demonstrate the case c). In contrast, in the case a), a single algorithm $\omega$ is generated to serve as $\omega_i^t$ in Eq.~(\ref{eq: ag}). In case b), a problem-specific $\omega_i$ is generated to serve as $\omega_i^t$ for each optimization step in solving $f_i$.

\subsubsection{\textbf{Related Works}}
Creating a comprehensive algorithm representation space $\Omega$ is crucial for the meta-level policy to produce innovative and efficient BBO algorithms. Current MetaBBO methodologies for AG can be categorized into three types based on their formulation of algorithm representation space $\Omega$: algorithm workflow composition, mathematical expressions, or natural/programming languages.

First, we introduce the works that perform algorithm workflow composition. A very early-stage work is conducted by Schmidhuber~\cite{meta-1} in 1987, where GP components are represented by the computer program space. Following such idea, GP is further applied to create improved EA variation operators~\cite{GP-AG-1,GP-AG-2}, evolve EA selection heuristics~\cite{GP-AG-3} and generate complete algorithm template~\cite{GP-AG-4}. At the meta-level, a GP is used to evolve low-level GP programs in a self-referential way. In the latest literature, GSF~\cite{yi2022GSF} first defines an algorithm template for EAs, then uses RL to fill each part of the template with operators from a predefined operator pool. 
ALDes~\cite{zhao2024aldes} overcomes the limitation of using fixed-length template through autoregresive learning. It first tokenizes the common algorithmic components and the corresponding configuration parameters in EAs, as well as the execution workflows such as loop and condition. Then, the algorithm generation task turns into a sequence generation task of the tokens. 
Concretely, ALDes prepares three types of operators: four ``selection for evolution'' operators, six evolution operators and five ``selection for replacement'' operators, each is associated with some hyper-parameters. Given the property of the target optimization problem, a Transformer-style policy auto-regressively selects a desired operator and configures its hyper-parameter from the candidate pool of each operator type. 


Second, we introduce the works that leverage mathematical expression to formulate $\Omega$. The motivation behind this line is that the design space of GSF and ALDes is highly dependent on manual engineering, which may limit the exploration of more novel algorithm structures. SYMBOL~\cite{2024symbol} addresses this issue by breaking down the update equations of BBO algorithms into atomic mathematical operators and operands. SYMBOL constructs a token set of common mathematical symbols used in EAs, such as $\{+,-,\times,x,x^*,x^-,x_i^*,\Delta x,x_r,c\}$. It then designs an LSTM-based policy which is capable of auto-regressively generating a sequence of these mathematical symbols. The generated sequence can be parsed into update equations for optimizing the low-level optimization problem. SYMBOL generates flexible update rules for each optimization step and each problem instance, bringing in certain self-adaptation capabilities. 

Third, we introduce works that leverage natural language and programming language to define $\Omega$. All works in this line leverage LLMs as their meta-level policies~\cite{ahmaditeshnizi2023optimus,2024llamoco,liu2023ael,liu2024eoh,2024llamea,huang2024llmopt}. The differences lie in the learning methodologies, the generation workflows and the target problem types. OptiMUS~\cite{ahmaditeshnizi2023optimus} leverages modular-structured LLM agents to formulate and solve~(mixed integer) linear programming problems. There are four agents in OptiMUS: formulator, programmer, evaluator, and manager, which constitute an optimization expert team and automate the algorithm generation task through their cooperation. 
To enable more general-purpose algorithm generation, AEL~\cite{liu2023ael} and EoH~\cite{liu2024eoh} are inspired by the evolution capability of large models~\cite{openelm}, 
prompting LLMs to perform mutation and crossover operations on code implementations of previous algorithms. 
After evolution, the best-so-far algorithm generated shows at most 24\% performance margin over human-crafted heuristics on Traveling Salesman Problems. 
Subsequent works such as LLaMEA~\cite{2024llamea} and LLMOpt~\cite{huang2024llmopt} generalize this paradigm to continuous BBO scenarios, and LLaMEA is shown to be capable of generating a more complex algorithm that is competitive with CMA-ES within 1\% score gap.
Despite the above works, LLaMoCo~\cite{2024llamoco} offers a novel perspective: instruction-tuning the general LLMs to act as an expert-level optimization programmer. LLaMoCo allows users to describe their specific optimization problems in Python/LaTex formulation, then it outputs the complete Python implementation of a desired optimizer for solving the given problems. To achieve this, a large-scale benchmarking is conducted to attain thousands of problem-solver pairs as the expert-level optimization knowledge. This knowledge is then injected into LLMs through instruction tuning. 
The experimental results in LLaMoCo demonstrates that a small model~(e.g., codeGen-350M) could generate superior algorithm program to larger models which are not fine-tuned by LLaMoCo (e.g., GPT-4), underscoring that domain specific knowledge might be the key for LLMs to understand, reasoning and solve problems.

\subsubsection{\textbf{Challenges}}
MetaBBO for AG works operate in a more expressive algorithm design space. The experimental results in some of these works demonstrate that the generated algorithms are on par with or even superior to human-crafted ones. 
The generated BBO algorithms can not only address optimization problems, but also be further analysed by human experts for novel insights in developing optimization techniques. 
Nevertheless, there are still several bottlenecks in existing works:
\begin{itemize}
    \item As an early-stage research avenue, related works in this area are still limited. More studies are expected to further unleash the potential of MetaBBO for AG.
    \item For symbolic system-based generation frameworks such as ALDes and SYMBOL, the token sets are relatively small, which leads to limited representation capability. How to construct a comprehensive and expressive token set tailored for BBO algorithm, and how to ensure the learning effectiveness in the enlarged algorithm design space need further investigation.
    \item For LLM-assisted MetaBBO for AG, although the workflow promises an efficient development pipeline, the computational resources required to obtain a competitive BBO algorithm are substantial. Besides, these works rely heavily on the prompt engineering, since LLMs are sensitive to the prompts they receive. 
\end{itemize}

\section{Different Learning Paradigms at Meta Level}\label{sec:4}
In this section, we introduce four key learning paradigms behind the majority of existing MetaBBO works: MetaBBO with reinforcement learning~(MetaBBO-RL), MetaBBO with supervised learning~(MetaBBO-SL), MetaBBO with neuroevolution~(MetaBBO-NE) and MetaBBO with in-context learning~(MetaBBO-ICL). 

\subsection{MetaBBO-RL}
In MetaBBO-RL, the meta-level algorithm design task is modeled as an MDP~\cite{1990mdp,sutton2018reinforcement}, where the environment is the low-level BBO process for a given problem instance $f$. The optimization state feature space for $s$, the algorithm design space $\Omega$, and the performance metric $\text{perf}(\cdot)$ serve as the MDP's state space, action space, and reward function, respectively. As a result, the meta-objective defined in Eq.~(\ref{eq:1}) becomes the expected accumulated reward. While various RL techniques can be applied, the choice must be made by considering the characteristics of the state and action spaces, which we categorize into three main types below.

\subsubsection{\textbf{Discrete State \& Discrete Action}} Tabular Q-learning~\cite{1992qlearning} and SARSA~\cite{sutton2018reinforcement} are value-based RL techniques that maintain a Q-table to iteratively update state-action values based on interactions with the environment. These methods have a notable benefit in their straightforward Q-table structures, which facilitates efficient convergence and reliable effectiveness. However, they are confined to MDPs with discrete (finite) state and action spaces. Many MetaBBO-RL works adopt these methods for their simplicity. In such works, optimization states and algorithm designs are pre-defined to form the rows and columns of the Q-table. At each optimization step $t$ in the low-level BBO process, the meta-level policy suggests an algorithm design $\omega^t$ according to $s^t$ and the Q table. Then, a transition $<s^t,\omega^t,\text{perf}(s^t, \omega^t, f),s^{t+1}>$ is obtained and the Q-table is updated as   

\begin{equation}
Q(s^t, \omega^t) = \text{perf}(s^t, \omega^t, f) + \gamma \max_{\omega \in \Omega} Q(s^{t+1}, \omega) 
\end{equation}
An example of this approach is the QLPSO algorithm~\cite{2020qlpso}, which dynamically adjusts the particle swarm topology. In QLPSO, the optimization states are $\{L2, L4, L8, L10\}$, representing different neighborhood size features of particles, with corresponding actions to either maintain or change the neighborhood size. Performance improvements resulting from successful topology adjustments are rewarded. Other works using similar methods include RLNS~\cite{2024PSORLNS},  QFA~\cite{2018QFA}, qlDE~\cite{huynh2021qlde}, RLMPSO~\cite{samma2016rlmpso},  DE-RLFR~\cite{2019derlfr}, QL-(S)M-OPSO~\cite{liu2019qlmopso}, MARLwCMA~\cite{sallam2020marlwcma}, LRMODE~\cite{huang2020lrmode}, RLEA-SSC~\cite{xia2021rlea-ssc},  RLDE~\cite{hu2021rlde}, RL-CORCO~\cite{hu2022rl-corco}, RL-SHADE~\cite{2022rlshade}.

\subsubsection{\textbf{Continuous State \& Discrete Action}} While Tabular Q-learning and SARSA are effective for discrete state spaces, some MetaBBO scenarios require continuous optimization states for finer algorithm design. In such cases, the MDP involves an infinite state space, making the Q-table structure incompatible. To address this, neural network-based Q-agents, such as DQN~\cite{2013dqn} and DDQN~\cite{2016ddqn}, are employed to handle continuous state features. The Q-agent is updated by minimizing the estimation error between the target and predicted Q-functions as
\begin{equation}\label{eq:dqn-metabborl}
\begin{split}
\text{Loss}(\theta)=\frac{1}{2}\biggl[ Q_\theta(s^t, \omega^t) -& \biggl(\text{perf}(s^t, \omega^t, f) \\+& \gamma \max_{\omega \in \Omega} Q_\theta(s^{t+1}, \omega)\biggr)\biggr]^2
\end{split}
\end{equation}
A representative example is DEDQN~\cite{2021dedqn}, where the optimization state is represented by four continuous FLA indicator features derived from a random walking strategy. The meta-level policy is an MLP Q-agent with three hidden layers. During the low-level BBO process, the Q-agent outputs Q-values for three candidate DE mutation operators and selects one for the current optimization step. The performance improvement after this step serves as a reward. The transition obtained is used to update the Q-agent by Eq.~(\ref{eq:dqn-metabborl}). Other works employing similar methods include R2-RLMOEA~\cite{tahernezhad2024r2-rlmoea}, DE-DDQN~\cite{2019deddqn}, MADAC~\cite{xue2022madac}, MOEA/D-DQN~\cite{tian2022modead-dqn}, CEDE-DRL~\cite{hu2023cede-drl}, SA-DQN-DE~\cite{liao2024sa-dqn-de}, UES-CMAES-RL~\cite{2024uescmaesrl}, HF~\cite{pei2024learning}.

\subsubsection{\textbf{Continuous State \& Continuous Action}} Building on the success of RL techniques in continuous control~\cite{2019rlrobotcontrol}, some MetaBBO-RL works adopt policy gradient-based methods (e.g., REINFORCE~\cite{1992REINFORCE}, A2C~\cite{1999Actor-critic}, PPO~\cite{2017ppo}) to handle both continuous states and algorithm designs. These allow for more flexible control of optimization behavior in the low-level BBO process, possibly improving performance. In this case, a policy neural network $\pi_\theta$ is used to output a probability distribution over the algorithm design space based on the optimization state. The gradient $\nabla_{\theta} J(\theta)$ used to update $\pi_\theta$ is computed as
\begin{equation}~\label{eq:pg-metabborl}
   \nabla_{\theta} J(\theta)=-\nabla_{\theta} \log \pi_{\theta}(\omega^t \mid s^t) \left(\sum_{t'=t}^{T} \gamma^{t'-t}\text{perf}(s^{t'}, \omega^{t'}, f)\right)
\end{equation}
We illustrate the method with the representative work GLEET~\cite{ma2024gleet}. In GLEET, the optimization state is represented by a structured feature set, including low-level information such as solution/objective space density and performance improvement indicators. A Transformer-style policy network (3 layers) outputs the posterior Gaussian distribution for each parameter of each individual. The concrete parameter values are then sampled from this distributions for the current optimization step, and the corresponding reward is assigned. After completing an optimization episode ($T$ steps), the policy network $\pi_\theta$ is updated by summing the gradients from each step, as shown in Eq. (\ref{eq:pg-metabborl}). Other MetaBBO-RL works employing similar methodologies include LTO~\cite{shala2020lto}, RLEPSO~\cite{yin2021rlepso}, LDE~\cite{sun2021lde}, RL-PSO~\cite{wu2022rl-pso}, MELBA~\cite{chaybouti2022melba}, MOEADRL~\cite{gao2023moeadrl}, LADE~\cite{liu2023lade}, RLAM~\cite{2023rlam}, AMODE-DRL~\cite{li2023amode-drl}, PG-DE~\cite{zhang2024pg-de}, GLEET~\cite{ma2024gleet}, RLEMMO~\cite{lian2024rlemmo}, RL-DAS~\cite{2024rldas}, SYMBOL~\cite{2024symbol}. %

\subsection{MetaBBO-NE}
Neuroevolution~\cite{1990ne} is a machine learning subfield where neural networks are evolved using EC methods rather than updated by gradient descent. In~\cite{2017ecrl}, ES is demonstrated as a scalable alternative to RL for MDPs, especially when actions have long-lasting effects. This inspired the development of MetaBBO methods using EC to evolve the policies, referred to as MetaBBO-NE. In MetaBBO-NE, the meta-level maintains a population of policies $\left\{\pi_{\theta_1},\dots,\pi_{\theta_K}\right\}$, with each policy $\pi_{\theta_k}$ being used to guide the algorithm design task for a training problem set. The fitness of each policy is the average performance gain across the problem instances in the training set. An EC method, such as ES, is employed to iteratively update the the meta-level policies, and after several generations, the optimal policy $\pi_{\theta^*}$ is obtained.

Representative works in MetaBBO-NE include LTO-POMDP~\cite{2021ltopomdp} and LGA~\cite{2023lga}. For example, in LGA, a population of attention-based neural networks is maintained at the meta-level, where each network functions as a neural GA to manipulate solutions. The OpenAI-ES~\cite{2017ecrl} is then used to evolve $K=32$ such networks over ten 10-dimensional synthetic functions from the COCO benchmark~\cite{coco}. 

\subsection{MetaBBO-SL}
MetaBBO works using the supervised learning paradigm are closely related to the meta task of solution manipulation~(SM). As we described in Eq.~(\ref{eq: ai}), SM aims to learn a parameterized meta-level policy $\pi_\theta$ as the low-level optimizer. The optimization process proceeds by iteratively calling $\pi_\theta$ to optimize the current (population of) solution(s). A key difference between MetaBBO-SL and MetaBBO-RL is that MetaBBO-SL meta-trains policies using direct gradient descent on an explicit supervising objective. This resembles regret minimization~\cite{zinkevich2007regret} of the target optimization problem’s objective function. To illustrate this, let us examine the recent work GLHF~\cite{2024glhf}, which proposes an end-to-end MetaBBO method mimicking a DE algorithm. GLHF unifies the DE mutation and crossover operations as matrix operations and designs $\pi_\theta$ as two customized network modules, LMM and LCM, to simulate matrix-based mutation and crossover. The Gumbel-Softmax function is used in the crossover module to make it differentiable. Given a solution population $X^t$ at the optimization step $t$ when optimizing a problem $f$, the $\pi_\theta$ in GLHF optimizes $X^t$ to generate offspring population: $X^{t+1} = \pi_\theta(X^t)$. The explicit supervising objective in this case is the objective value $f(X^{t+1})$, which serves as a regret function to minimize. Then the gradient used to update the policy at step $t$ is computed as
\begin{equation}
   \nabla_{\theta} J(\theta) \propto \frac{\partial f(X^{t+1})}{\partial \pi_\theta} \cdot \frac{\partial \pi_\theta}{\partial \theta}
\end{equation}
Minimizing this regret-based objective trains the meta-level policy for effective optimization on the target problem. However, the differentiability of $f$ is a requirement, which may not hold for ``black-box'' problems. 
Other works in this line include RNN-OI~\cite{2017rnnoi}, RNN-Opt~\cite{2019rnnopt}, B2Opt~\cite{li2023b2opt}, EvoTF~\cite{2024evotf}, LEO~\cite{brahmachary2024leo}, RIBBO~\cite{2024ribbo}, NAP~\cite{2024nap}. Notably, RIBBO~\cite{2024ribbo} and EvoTF~\cite{2024evotf} use supervised imitation learning to meta-train their policies to mimic a teacher BBO algorithm. For instance, RIBBO uses a GPT architecture to imitate optimization trajectory from diverse existing BBO algorithms. It tokenizes each of the collected optimization trajectories into a target token sequence $\{R^1,X^1,Y^1,...,R^T,X^t,Y^T\}$, where $R^t$, $X^t$ and $Y^t$ are the regret-based explicit supervising objective, population positions and objective values respectively. 
Through behavior cloning, RIBBO trains the GPT to mimic these trajectories, resulting in a generalized optimization behavior and outperforming baselines on 3 out of 5 problems.

\renewcommand{\arraystretch}{1.5} 

\begin{figure*}[t]
    \centering
    \includegraphics[width=0.93\linewidth]{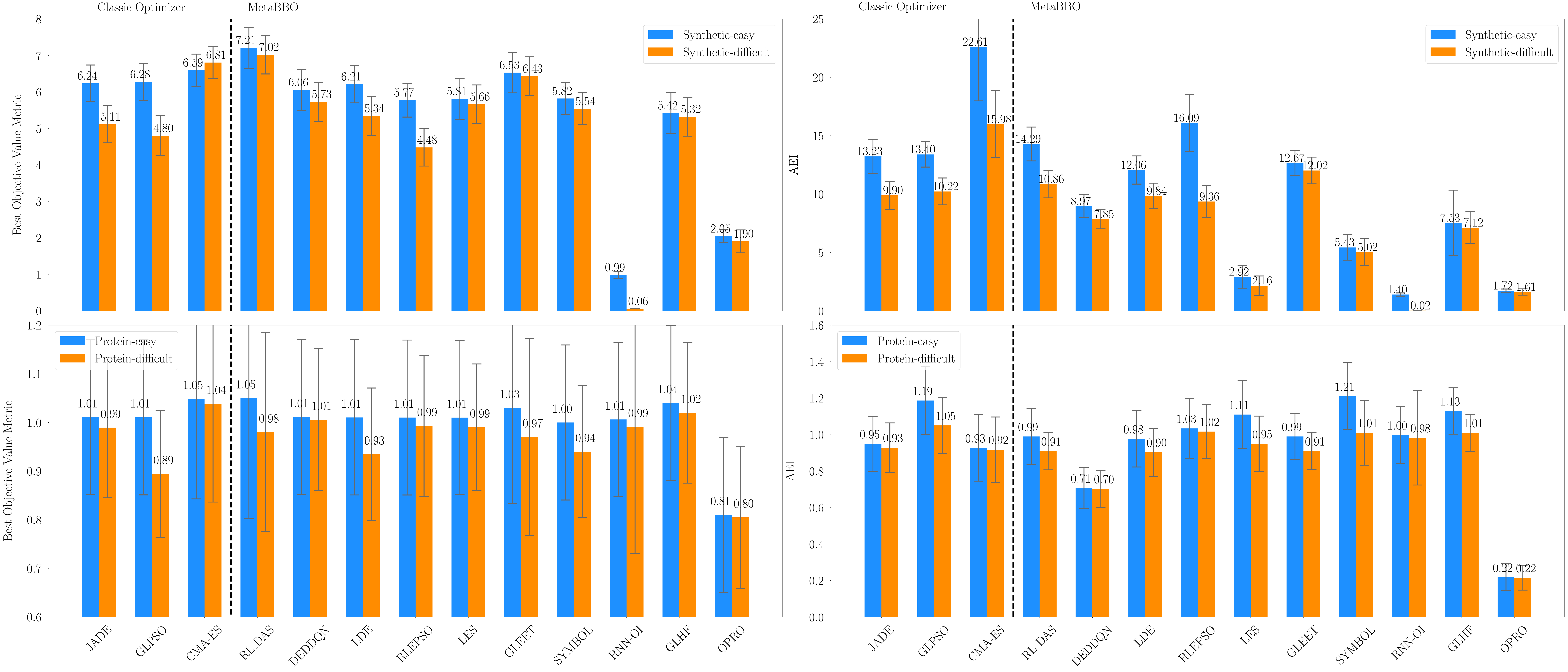}
    \caption{Performance comparisons. \textbf{Top Left:} Best objective values on synthetic testsuites. \textbf{Top Right:} AEI scores on synthetic testsuites. \textbf{Bottom Left:} Best objective values on protein docking testsuites. \textbf{Bottom Right:} AEI scores on protein docking testsuites.}
    \label{fig:AEI_Obj_synthetic_and_protein}
\end{figure*}

\subsection{MetaBBO-ICL}
In-Context Learning (ICL)~\cite{ICL} is a popular paradigm in LLM research, which prompts LLMs with a structured text collection: a \emph{task description}, several \emph{in-context examples}, and a concrete \emph{task instruction}. This structured prompt enables LLMs to reason effectively based on the provided context, without requiring gradient descent or parameter updates. MetaBBO-ICL is closely related to two meta tasks: Solution Manipulation (SM) and Algorithm Generation (AG). The main distinction between existing works lies in how they construct effective in-context prompts. 

For the SM task, OPRO~\cite{yang2024opro} introduces optimization via iterative prompting. In each iteration, the \emph{task description} is tailored to the specific problem, including the objective function and constraints in natural language. The \emph{in-context examples} consist of prior optimization trajectories, and the \emph{task instruction} asks the LLM to find a solution better than the previous best. However, this approach faces challenges due to the limited optimization expertise of general LLMs, which are not typically trained with optimization knowledge in mind~\cite{2024llamoco}. Recent studies creatively suggest guiding LLMs to mimic certain EAs~\cite{liu2024lmea}, which involves directing LLMs to execute mutation, crossover, and elitism strategies on the specified \emph{in-context examples}. 

For the AG task, the core idea is using LLMs to understand and evolve optimizer programs. Note that evolving programs is not a novel concept. This topic traces back to Genetic Programming~(GP) method, which performs evolution of computer program within the code space in a self-referential way: evolve evolution algorithms. Leveraging the semantic reasoning ability of CodeLLMs for program evolution, initial works such as Funsearch~\cite{funsearch} and EUREKA~\cite{ma2023eureka} discover competitive heuristic program and reward design respectively.  Following these works, in MetaBBO-ICL, researchers begin to discuss the possibility of evolving competitive optimization program with CodeLLMs. A representative work in this area is EoH~\cite{liu2024eoh}, where the \emph{task description} includes both the optimization problem formulation and a concrete algorithm design task. The LLM is asked to first describe a new heuristic and then implement it in Python. The \emph{in-context examples} consist of previously suggested programs, while the \emph{task instruction} provides five evolution instructions, each with varying levels of code refinement. Other related works include AEL~\cite{liu2023ael}, LLaMEA~\cite{2024llamea}, and LLMOPT~\cite{huang2024llmopt}.

In this section, we have introduced four key learning paradigms widely adopted in existing MetaBBO works. To summarize, each learning paradigm has its own advantages and downsides. By the aid of powerful LLMs, MetaBBO-ICL spends least efforts in developing and designing. In contrast, MetaBBO-RL requires certain expertise in designing effective RL systems, while achieving significant policy improvement. As for the final performance, it is empirically observed that MetaBBO-RL and MetaBBO-SL generally achieve superior optimization results~(see Fig.~\ref{fig:AEI_Obj_synthetic_and_protein}). As for the research interests for the four learning paradigms, we found MetaBBO-RL is continuously catching researchers' attention, while MetaBBO-ICL rises quickly~(see Table~\ref{tab:low-level}).

\section{Empirical Evaluation}\label{sec:5}
\subsection{Development in Benchmarks}

For benchmarking BBO optimizers, many well-known testsuites have been extensively studied and developed~\cite{cec2021,coco2019}. With the ongoing development of BBO, the corresponding benchmarks aim to 1) propose more diverse benchmark problems in synthetic~\cite{moo-bench,mmo-bench,do-bench,lsgo-bench,tianye-bench,evobbo,ma-bbob} and realistic~\cite{aclib,ne-bench,realistic-bench} scenarios; and 2) automate the benchmarking process through a software platform~\cite{doerr2018iohprofiler,coco}. These traditional BBO benchmarks can serve as evaluation tools for MetaBBO methods. However, compared with traditional EC algorithms, the system structure of MetaBBO is more intricate. Its bi-level learning paradigm involves a meta-level policy, a low-level optimizer, the training/testing logic of the entire system, and the interfaces between the meta and lower levels. This complexity creates a gap between the conventional BBO benchmarks and MetaBBO methods. To address this compatibility issue, a recent work termed  MetaBox~\cite{2024metabox} proposes the first benchmark platform specifically for developing and evaluating MetaBBO methods. It provides three different single-objective numerical problem collections~(Synthetic-10D, Noisy-Synthetic-10D, Protein-Docking-12D), along with two different train-test split modes~(easy and difficult), which benefits MetaBBO's training under different problem distributions and difficulties. In the next subsection, we provide a proof-of-principle evaluation of several representative MetaBBO methods using MetaBox.

\subsection{Proof-of-Principle Evaluation by MetaBox}
In this section, we use MetaBox~\cite{2024metabox} to evaluate the performance of three traditional EC algorithms and ten representative MetaBBO methods. For traditional EC algorithms, we empirically select three representative algorithms: JADE~\cite{2009jade}, GLPSO~\cite{2015glpso} and CMA-ES~\cite{hansen2016cmaes} from three mainstream traditional BBO classes: DE~\cite{1997de}, PSO~\cite{1995pso} and ES~\cite{2002es}, respectively. 
Further, we include the MetaBBO methods covering all four meta-tasks and all four learning paradigms: 
\begin{itemize}
    \item \textbf{AS:} RL-DAS~\cite{2024rldas} (RL-based dynamic selection);
    \item \textbf{AC:} DE-DDQN~\cite{2019deddqn} (RL-based AOS); LDE~\cite{sun2021lde}/ RLEPSO~\cite{yin2021rlepso}/GLEET~\cite{ma2024gleet} (RL-based HPO); LES~\cite{2023les} (NE-based HPO);  
    \item \textbf{AG:} SYMBOL~\cite{2024symbol} (RL-based symbolic synthesis); 
    \item \textbf{SM:}  RNN-OI~\cite{2017rnnoi}/GLHF~\cite{2024glhf} (early/recent SL-based trajectory prediction); OPRO~\cite{yang2024opro} (ICL-based linguistic optimization).
\end{itemize}
All experiments follows the protocols in MetaBox. Due to the space limitation We leave the detailed baseline selection criteria and experimental setup in Appendix I.A \& I.B respectively.

The AEI score in MetaBox~\cite{2024metabox} evaluates the overall optimization performance of a MetaBBO method by aggregating three key metrics: final optimization results, FEs consumed, and runtime complexity, using an exponential average, lager is better. The left side of Fig.~\ref{fig:AEI_Obj_synthetic_and_protein} presents the final optimization accuracy of all baselines on Synthetic BBOB (top) and Realistic Protein Docking (bottom) testsuites, while the right side presents their respective AEI scores. The results show that: 
1) When considering only the final accuracy, MetaBBO methods such as RL-DAS, LDE, and GLEET achieve comparable or even superior performance to traditional BBO optimizers, while some other MetaBBO methods still perform inferiorly compared to traditional BBO methods. This indicates that while MetaBBO methods show potential, as an emerging topic, there is still significant room for improvement. 2) Different evaluation metrics yield different conclusions regarding performance. When considering both optimization performance and computational overhead, traditional BBO optimizers such as CMA-ES achieve a significantly better trade-off, as shown on the right side of Fig.~\ref{fig:AEI_Obj_synthetic_and_protein}. This highlights a potential limitation of MetaBBO methods: they typically involve additional computation during the meta-level decision-making process.
3)~Comparing the performance on the Synthetic BBOB and Protein Docking test suites, we observe that the performance gap between MetaBBO methods and traditional BBO optimizers narrows on both evaluation metrics as the problem type shifts from the relatively simpler synthetic set to the more challenging realistic protein docking set. This suggests that MetaBBO is promising for solving complex optimization problems.
4) MetaBBO-RL methods~(including RL-DAS, DE-DDQN, LDE, RLEPSO, SYMBOL) outperform MetaBBO-NE methods~(LES), MetaBBO-SL methods~(RNN-OI) and MetaBBO-ICL methods~(OPRO). 
This observation highlights an important future direction for the MetaBBO domain: analyzing the theoretical performance bounds of different MetaBBO methods. 5) The AEI of OPRO~(MetaBBO-ICL method) is significantly lower, this might indicate that iteratively optimization paradigm through in-context prompting LLMs is severely challenged by the efficiency issue.

Besides the above algorithmic performances, one should also examine a MetaBBO method's learning capability. As a learning system, it is expected that a MetaBBO method should show certain generalization ability on unseen problem instances/distributions. To this end, we have tested the MetaBBO baselines on MetaBox for their Meta Generalization Decay~(MGD) Meta Transfer Efficiency~(MTE), two indicators proposed in MetaBOx to measure a MetaBBO method's learning ability. Detailed results are provided in Appendix II.

\begin{figure*}[t]
    \centering
    \includegraphics[width=0.99\linewidth]{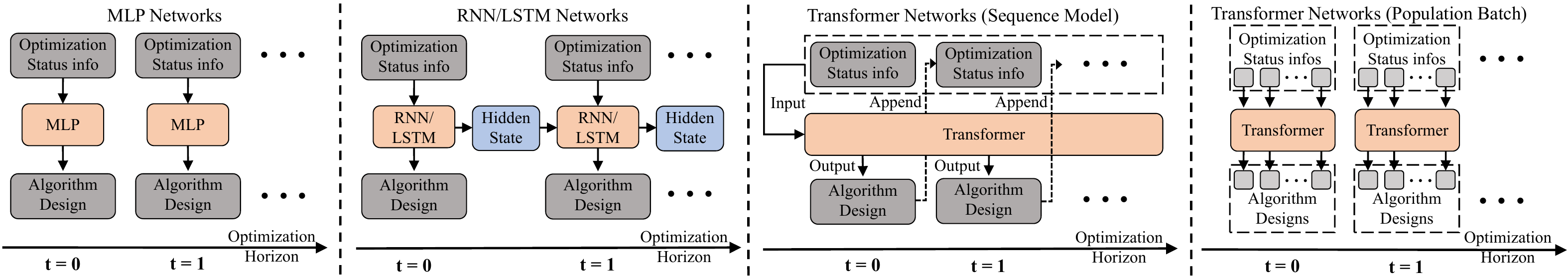}
    \vspace{-1mm}
    \caption{The workflow of different neural networks used in existing MetaBBO works: MLP, RNN/LSTM and Transformer architectures.}
    \label{fig:networkdesign}
    \vspace{-3mm}
\end{figure*}

\section{Key Design Strategies}\label{sec:6}

\subsection{Neural Network Design}\label{sec:NNdesign}
Four common neural network architectures are frequently adopted: 1) MLP, 2) RNN and LSTM, 3) temporal dependency Transformer, and 4) spatial dependency Transformer, as illustrated in Fig.~\ref{fig:networkdesign} from left to right. The basic MLP (leftmost in Fig.~\ref{fig:networkdesign}) is widely used in existing works due to its simplicity and efficiency in training and inference. However, the MLP is limited in analyzing the temporal and data batch dependencies within the low-level BBO process. We next introduce novel designs that address these limitations.

\subsubsection{\textbf{Temporal Dependency Architectures}} 
The low-level BBO process involves iterative optimization over $T$ generations. A basic MLP-based policy may struggle to effectively leverage historical information along the optimization trajectory. Then, an intuitive solution is to introduce architectures that support temporal sequence modeling. To this end, works such as RNN-OI~\cite{2017rnnoi}, RNN-Opt~\cite{2019rnnopt}, and LTO-POMDP~\cite{2021ltopomdp} introduce RNNs and LSTMs~\cite{hochreiter1997lstm}, which integrate historical optimization information into hidden representations and combine it with the current optimization state (shown in the second part of Fig.~\ref{fig:networkdesign}). While these approaches improve learning effectiveness by incorporating historical information, training on long horizons (often involving hundreds of generations) using RNN/LSTM can be challenging due to the inherent issues of gradient vanishing or explosion. Subsequent works, such as MELBA\cite{chaybouti2022melba}, RIBBO~\cite{2024ribbo}, and EvoTF~\cite{2024evotf}, address this limitation by leveraging Transformer architectures for better long-sequence modeling. The common workflow in these works is illustrated in the third part of Fig.~\ref{fig:networkdesign}, where a trajectory of historical optimization states is processed by the Transformer to inform the next step in algorithm design.

\subsubsection{\textbf{Spatial Dependency Architectures}} 
In addition to temporal properties, a key characteristic of EC is its population-based search manner. Recent MetaBBO methods tailor algorithmic components for each individual in the population, maximizing flexibility for low-level optimization. As illustrated in the rightmost part of Fig.~\ref{fig:networkdesign}, works such as LGA\cite{2023lga}, LES~\cite{2023les}, B2Opt~\cite{li2023b2opt}, GLEET~\cite{ma2024gleet}, RLEMMO~\cite{lian2024rlemmo}, and GLHF~\cite{2024glhf} construct optimization state features as a collection of individual optimization states and leverage the Transformer's attention mechanism to enhance information sharing across the population of individuals. For instance, GLEET~\cite{ma2024gleet} proposes a novel Transformer-style network that includes a ``fully informed encoder'' and an ``exploration-exploitation decoder''. The encoder promotes information sharing by applying self-attention to the state features of all individuals. The decoder then decodes the hyperparameter values for each individual specifically. Besides mining the spatial dependency in solution space, a recent work termed as TabPFN~\cite{TabPFN} leverages attention mechanism and bayesian prior to discover the spatial dependency in problem instance space, which improves the performance in per-instance algorithm selection~\cite{ana_kostovska_2023_7807853}.


\subsection{State Feature Design}\label{sec:feature}
A key component for MetaBBO’s generalization across diverse optimization problems is the state feature extraction function sf$(\cdot)$. We identify three types of features:
a)~\emph{Problem identification features}, which captures the landscape properties of the target problem.
b)~\emph{Population profiling features}, which describes the distribution of solutions in the low-level BBO process.
c)~\emph{Optimization progress features}, which tracks improvements in the solution evaluations at each step.
Next, we introduce common practices for preparing these features.

\subsubsection{\textbf{Problem Identification Features}} 
To identify the target optimization problem, the Exploratory Landscape Analysis (ELA) framework~\cite{2011ela} is widely used for single-objective optimization problems. ELA includes six groups of metrics, such as local search, skewness of the objective space, and approximated curvature (both first and second-order), which provide a comprehensive summary of the problem's landscape properties. To compute ELA features, a large number of points are sampled from the BBO problem and used to compute the features. For example, linear and quadratic models are fitted to the sampled points and their objective values, and the resulting model parameters have been shown to be useful for differentiating different problems. For multi-objective optimization, the features can be obtained by decomposing the problem into single-objective sub-problems and conducting single-objective feature analysis techniques~\cite{moo-landscape-sv1}.

\subsubsection{\textbf{Population Profiling Features}} 
In an optimization problem, the solution population can converge to different regions of the fitness landscape. MetaBBO aims to dynamically adapt algorithm designs to help the low-level optimizer adjust to these diverse regions. To analyze the distribution of the population, Fitness Landscape Analysis (FLA)\cite{fla} is commonly used, providing various indicators, such as Fitness Distance Correlation\cite{fdc}, Ruggedness of Information Entropy~\cite{rie}, Auto-Correlation Function~\cite{acf}, Dispersion~\cite{dispersion}, Negative Slope Coefficient~\cite{nsc}, and Average Neutral Ratio~\cite{anr}. Some of these indicators measure local landscape properties based on the population’s location in the fitness space. 
Population profiling features complement problem identification features, providing the meta-level policy with a more accurate optimization state for specific optimization steps.

\subsubsection{\textbf{Optimization Progress Features}} 
Optimization progress features further complement ELA and FLA features by providing the meta-level policy with additional information on objective evaluation-related properties, such as the consumed FEs, and the distribution of the current population along with the objective values. These features track the improvement and convergence of the population. Interestingly, recent MetaBBO works like DE-DDQN~\cite{2019deddqn}, RLEPSO~\cite{yin2021rlepso}, and GLEET~\cite{ma2024gleet} have found that optimization progress features alone can be sufficient for learning a generalizable meta-level policy. A key reason is that computing ELA/FLA features consumes additional FEs, which reduces the learning steps available for the meta-level policy, thus degrading both learning effectiveness and final optimization performance.

\subsection{Training Distribution Design}\label{sec:traindist}
The training problem set is crucial for learning a generalizable meta-level policy, with diversity being a key factor. Early works like RNN-OI~\cite{2017rnnoi} were trained on a limited set of instances from the CoCo-BBOB test suite. As shown in Fig.~\ref{fig:AEI_Obj_synthetic_and_protein}, a narrow training set leads to poor generalization. To enhance the diversity of the training set, two main methodologies are commonly used in existing MetaBBO approaches.

\subsubsection{\textbf{Augmenting Existing Benchmarks}}
 Standard BBO benchmarks include the CoCo-BBOB~\cite{coco2010,coco2019} and CEC BBOB-Competition~\cite{cec2005,cec2021} testsuites, which contain approximately 20–30 synthetic functions with various properties like multimodality, non-separability, and non-convexity. Most MetaBBO works augment these testsuites by mathematical transformations: given a D-dimensional function instance $f(x): \mathbb{R}^D \rightarrow \mathbb{R}$, it can be transformed to a new instance $f^\prime(x) = f(M^T(x - o))$, where $M \in \mathbb{R}^{D\times D}$ is a rotation matrix and $o \in \mathbb{R}^D$ is an offset to the optimal. For example, recent works like GLEET~\cite{ma2024gleet} and RL-DAS~\cite{2024rldas} apply random combinations of shifts and rotations on the CEC2021 test suite~\cite{cec2021}, generating thousands of synthetic instances and significantly improving the generalization performance of the learned meta-level policy.

\subsubsection{\textbf{Constructing New Benchmarks}}
\begin{figure}[t]
    \centering
    \includegraphics[width=0.85\columnwidth]{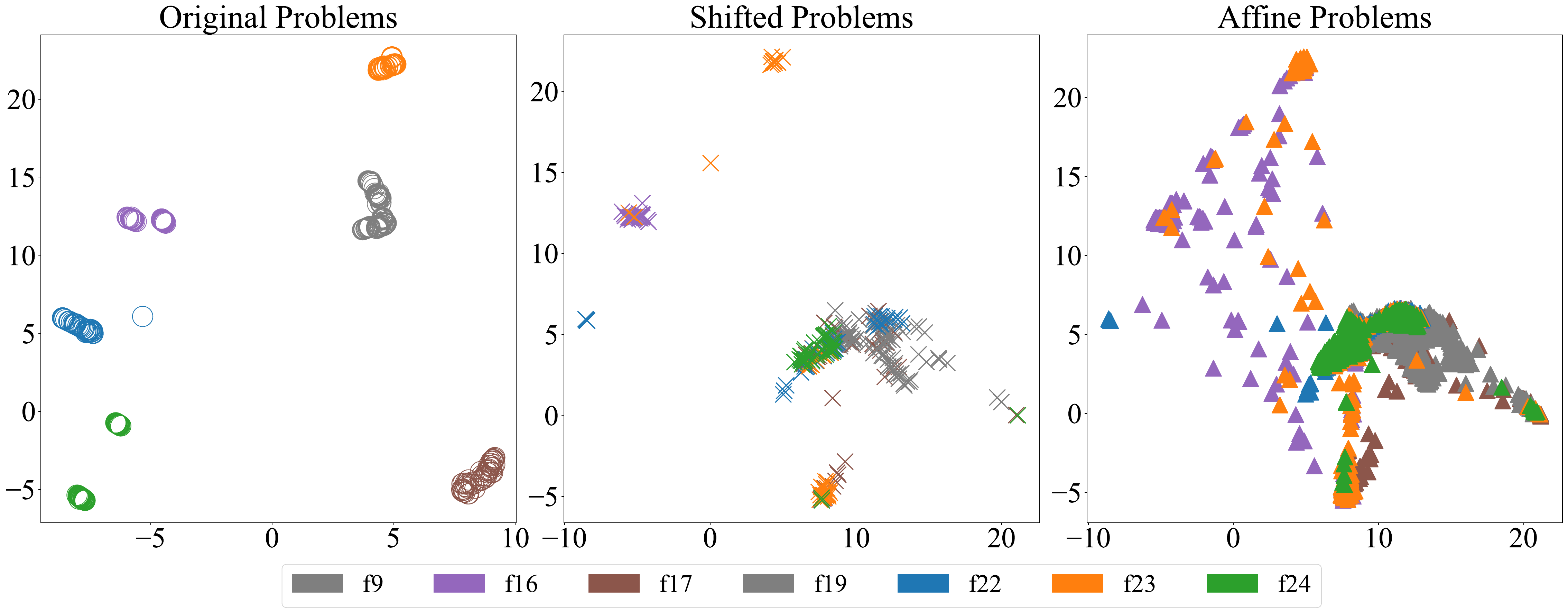}
    \vspace{-1mm}
    \caption{The projected 2D ELA distributions of \textbf{Left:} the original BBOB problems; \textbf{Middle:} the BBOB problems with shifted optimum; \textbf{Right:} the BBOB problems generated by MA-BBOB~\cite{ma-bbob}. }
    \label{fig:ELA_projection}
    \vspace{-3mm}
\end{figure}

While augmenting existing standard synthetic functions with shift and rotation transformations improves generalization, there is still room for greater diversity in the problem set. To illustrate this, we show the 2D projection of the ELA distribution for some CoCo-BBOB problem instances and their transformed counterparts in the left and middle of Fig.~\ref{fig:ELA_projection}. The results reveal that the transformations introduce some diversity
, but the improvement is still limited. 
There also exist a few studies that focus on the sensitivity of the landscape features to represent the benchmark functions, such as Škvorc et al.~\cite{vskvorc2021effect} reveal the importance of sampling methods in the invariance of landscape features, and Prager et al.~\cite{prager2023nullifying} analyze the the sensitivity of landscape features to absolute objective values.
More effective approaches are expected to generate novel benchmarks. The recent work MA-BBOB\cite{ma-bbob} demonstrates that affine combinations of existing synthetic functions can create more diverse instances. This is shown in the right part of Fig.~\ref{fig:ELA_projection}, where the instances generated by MA-BBOB covers wider feature space.

\subsection{Meta-Objective Design}\label{sec:objective}
The meta-objective $J(\theta)$ in MetaBBO represents the expected accumulated performance gain $\text{perf}(\cdot)$ over the problems in the training set. In existing MetaBBO works, $\text{perf}(\cdot)$ is typically tied to the objective values of the solution population, guiding the meta-level policy toward improved optimization performance. An intuitive approach is to use an indicator function: if performance improves between two optimization steps, a positive reward is given; otherwise, a negative or zero reward is assigned. This approach is widely used in early MetaBBO works such as  DE-DDQN~\cite{2019deddqn}, QLPSO~\cite{2020qlpso}, and MARLwCMA~\cite{sallam2020marlwcma}. 

\subsubsection{\textbf{Scale Normalization}}
Nevertheless, this basic approach can pose challenges when aiming to precisely assess performance improvements, which in turn could affect the adaptability of the learned policy. An alternative method involves computing $\text{perf}(\cdot)$ directly by determining the reduction in the objective value, expressed as $\Delta f^t = f^{*,t-1} - f^{*,t}$. However, directly using this absolute objective value descent may lead to unstable learning due to differing objective value scales across various optimization problems. To mitigate this issue, recent MetaBBO works apply normalization to the objective descent:
\begin{equation}\label{eq:norm-reward}
    \text{perf}(\cdot,t) = \frac{f^{*,t-1} - f^{*,t}}{f^{*,1} - f^*}
\end{equation}
where $f^{*, 1}$ denotes the objective value of the best solution in the initialized population, and $f^*$ represents the optimum of $f$. In practice, $f^*$ is unknown because $f$ is a black-box function. However, it can be approximated by an efficient BBO algorithm running in advance.

\subsubsection{\textbf{Sparse Reward Handling}}
The difficulty of the low-level BBO process increases over time. Initially, the objective value may decrease rapidly, but later, the descent slows as convergence approaches, often resulting in a sparse reward issue in learning systems. This can mislead the learning of the meta-level policy, causing it to favor sub-optimal algorithm designs that focus primarily on the early stages of optimization. To address this, recent works introduce an adaptive performance metric with a scale factor $\lambda(t)$ to Eq.~(\ref{eq:norm-reward}) to amplify the performance improvement in the later optimization stages:
\begin{equation}\label{eq:norm-reward-add}
    \text{perf}(\cdot,t) = \lambda(t) \times \frac{f^{*,t-1} - f^{*,t}}{f^{*,1} - f^*}
\end{equation}
where the scale factor $\lambda(t)$ is a incremental function of the optimization step $t$. For instance, in MADAC~\cite{xue2022madac}, $\lambda(t)$ is $2f^*-f^{*,t-1}-f^{*,t}$. However, ablation studies in RL-DAS~\cite{2024rldas}, GLEET~\cite{ma2024gleet}, RIBBO~\cite{2024ribbo}, and GLHF~\cite{2024glhf} suggest that using the unscaled, exact performance improvement metric without a scaling factor may be more effective. This underscores the variability in scaling methods' effectiveness across different MetaBBO tasks, warranting further investigation.

\section{Vision for the Field}\label{sec:7}
\subsection{Generalization Toward Task Mixtures}
A promising direction is the generalization toward a mixture of tasks. While MetaBBO works have explored various aspects of model generalization, the evaluation and analysis we provide in previous sections outline potential improvement through advanced learning techniques, e.g., transfer learning \cite{2016transfer-learning-survey} and multitask learning \cite{2021multitask-learning-survey}. 

First, existing works often focus on algorithm design for specific optimizers. For instance, methods like LDE \cite{sun2021lde} and GLHF \cite{2024glhf} are designed for algorithm configuration or imitation tasks, but primarily with basic DE. This narrow focus might lead to uncertain performance when applying these methods to other optimizers. A more effective approach would be to create a higher-level framework that defines MetaBBO tasks across multiple optimizers, establishing a multitask design space. Developing a universal modularization paradigm for various optimizers could allow training a meta-level policy that generalizes well across tasks.

Additionally, existing works focus exclusively on a single specific problem type. Separate policies are trained for each task type, leading to increased complexity. This outlines an opportunity to develop a unified agent capable of engaging in automatic algorithm design that adapts to various problem types. This method not only streamlines the optimization process but also aligns more closely with real-world scenarios, where practitioners frequently encounter a diverse array of problems. To overcome the limitations of existing methods, a universal problem representation system is essential to bolster the generalization across diverse problem domains.

\subsection{Fully End-to-End Autonomy}
The main motivation behind MetaBBO is to reduce the labor-intensive need for expert consultation by offering a general optimization framework. However, existing MetaBBO approaches still introduce design elements that rely on expert knowledge to enhance performance. This reliance typically involves: 1) the low-level optimization state $s$, often hand-crafted as a feature vector to represent problem properties or optimization progress; and 2) the meta-objective, which is mostly developer-defined, introducing subjectivity. While initial efforts have been made to automate feature extraction using neural networks~\cite{deep-ela,neurELA,rldeafl} and employ model-based RL to learn the meta-objective objectively~\cite{zhao2023marlabc}, further systematic studies are needed. Besides, MetaBBO focuses on designing algorithms in isolation, assuming the optimization problem is predefined and ready for evaluation. In reality, the initial step often involves formulating the problem, either through manual model construction~\cite{ma-bbob} or data-driven methods~\cite{clddea,ccddea,sddobench,surrrlde}. This disconnect reveals a major gap in the optimization process. A more integrated approach would involve objective formulation learning, automatic feature extraction and customized algorithm design. Developing a cohesive pipeline for these steps offers a promising direction for advancing optimization and improving problem-solving in practical applications.

\subsection{Smarter Integration of LLMs}
Although existing MetaBBO-ICL methods have shown possibility of leveraging general LLMs to assist algorithm design tasks, they still face challenges considering computational efficiency, code quality, intepretability, quality \& diversity control and reasoning ability for complex optimization tasks. It has been observed MetaBBO-ICL methods~(e.g., OPRO~\cite{yang2024opro} and LMEA~\cite{liu2024lmea}) have to consume 100-200k tokens during the iterative in-context conversation with LLMs to achieve certain optimization performance~\cite{2024llamoco}. Such efficiency issue is addressed by instruction-tuning general LLMs in ~\cite{2024llamoco}. However, the results in this work show that the error rate of the output code is 5\% - 10\%, which degrades the final performance. Considering the intepretability, since works such as OPRO implicitly prompt LLMs to learn the intent for optimization performance improvement, the logic behind the LLMs reasoning workflow is still black-box to us, which hinders the feedback loop between LLMs and human experts. Besides, existing MetaBBO-ICL methods merely focus on the quality \& diversity control within the optimization process. Future works must allow explicit commands for such exploration-exploitation tradeoff~\cite{song2024position}. Last but not least, recent researches such as~\cite{mirzadeh2025gsmsymbolic, zhang2024training} indicate the fragility of mathematical reasoning in existing LLMs, which further questions those LLM-assisted MetaBBO approaches.

This suggests two promising directions: First is the automated MetaBBO workflow search, leveraging LLMs for designing MetaBBO workflow through code generation and function search. Designing a learning system like MetaBBO is inherently challenging, as it requires considerable expertise. By providing LLMs with foundational principles of MetaBBO, the chain of thought within the models may uncover novel paradigms. 
Second, enhancing the semantic understanding of LLMs regrading optimization processes, terminologies, programming logics, problem descriptions would significantly elevate their expertise. To achieve this, an interesting direction is to develop symbolic language tailored to optimization domain, establishing a comprehensive grammar system and accumulating sufficient use cases to train a foundation model specifically for optimization. Third, since LLM-based MetaBBO approaches implicitly requires (massive) pre-training, an imminent future work for MetaBBO is to construct fair benchmarking standards and platforms that could compare MetaBBO approaches with traditional BBO methods fairly in both computational cost and optimization performance.

\section{Conclusion}
In this survey, we provide a comprehensive review of recent advancements in MetaBBO. As a novel research avenue within the BBO and EC communities, MetaBBO offers a promising paradigm for automated algorithm design. Through a bi-level data-driven learning framework, MetaBBO is capable of meta-learning effective neural network-based meta-level policies. These policies assist with algorithm selection and algorithm configuration for a given low-level optimizer, as well as to imitate or generate optimizers with certain flexibility. 

Our review begins with the mathematical definition of MetaBBO, clarifying its bi-level control workflow. Next, we systematically explore four main algorithm design tasks where MetaBBO excels: AS, AC, SM, and AG. Following the discussion of these tasks, we examine four methodologies of training MetaBBO: SL, RL, NE, and ICL. We hope these two parts will provide readers with a clear roadmap to quickly locate their interested MetaBBO methods. Furthermore, we provide comprehensive benchmarking on latest MetaBBO approaches considering their computational efficiency, optimization performance and learning capability, revealing current works' significance and limitations. Subsequent in-depth analysis on some core designs of MetaBBO: the neural network architecture, optimization state feature extraction mechanism, training problem distribution, and meta-objective design. These insights offer practical guidelines for researchers and practitioners aiming to develop more effective and efficient MetaBBO methods. At last, we propose several interesting and open-ended future directions for MetaBBO research, encouraging further exploration and innovation in this promising field. 


\bibliographystyle{IEEEtran}
\bibliography{ref}
\newpage
\onecolumn
\appendix
\section{Experimental Settings}
\subsection{Baseline Selection Criteria}
We conduct the proof-of-principle evaluation of several representative BBO and MetaBBO methods to provide community some insights of MetaBBO method and metric designs. Therefore we select baselines in the empirical evaluation showcase following two motivations:
\begin{itemize}
    \item Since a large number of MetaBBO methods focus on bound-constrained single-objective numerical optimization and existing MetaBBO methods are not capable across different problem types, we evaluate traditional BBO and MetaBBO methods on single-objective numerical optimization testsuites as a showcase.
    \item To conduct a comprehensive evaluation, we attempt to cover all categories summarized in the paper, including different meta tasks (AS, AC, SM and AG), different learning paradigm (RL, SL and NE) and different low-level BBO algorithm classes (i.e., DE, PSO and ES). We did not include MetaBBO-ICL methods with LLMs because of their high time cost and API fee. But we notice the concerns raised by reviewers about the in-depth analysis on LLM-aided MetaBBO, therefore we add a classic MetaBBO-ICL method, OPRO~\cite{yang2024opro}, in the comparison in Section V, Part B, which shows poor performance due to the weak optimization embedded in LLM and the long LLM conversation time cost. 
\end{itemize}

Following the two motivations above, we adopted a recent single-objective numerical optimization benchmark MetaBox~\cite{2024metabox} to evaluate the performance traditional EC algorithms and MetaBBO methods. For traditional EC algorithms, we select algorithms from three mainstream EC classes for BBO: DE~\cite{1997de}, PSO~\cite{1995pso} and ES~\cite{2002es}. For each of the classes, we choose one representative, highly cited and advanced algorithm as the representative, therefore for DE we select JADE~\cite{2009jade}, for PSO we choose GLPSO~\cite{2015glpso} and for ES we integrate CMA-ES~\cite{hansen2016cmaes}. Further, we include recent and advanced MetaBBO methods covering all four meta-tasks: for algorithm selection we adopt a recent RL-based method RL-DAS~\cite{2024rldas}. For algorithm configuration, we include the RL-based robust operator selection method DE-DDQN~\cite{2019deddqn} in MetaBox. Three hyper-parameter optimization MetaBBO methods, RL-based LDE~\cite{sun2021lde}, RLEPSO~\cite{yin2021rlepso} and NE-based LES~\cite{2023les}, are included covering the three mainstream EC classes DE, PSO and ES respectively. Besides, a recent advanced hyper-parameter optimization framework GLEET~\cite{ma2024gleet} which could compatibly control both DE and PSO is also selected. For algorithm generation, a recent MetaBBO-RL method SYMBOL~\cite{2024symbol} is included. For solution manipulation, we choose two MetaBBO-SL methods: the earliest work RNN-OI~\cite{2017rnnoi} and a recent method GLHF~\cite{2024glhf}. For all baselines above we use the implementations integrated in MetaBox. Besides, OPRO~\cite{yang2024opro} as a MetaBBO-ICL solution manipulation method is also included in the revised paper.

\subsection{Experimental Setup}
In the paper, all baselines are tested on three single-objective numerical testsuites: 10D Synthetic Testsuite with 24 problem instances, 10D Noisy Synthetic Testsuite with 24 problem instances and 12D Protein-Docking Testsuite with 280 problem instances. The ``easy'' postfix indicates that 75\% problem instances are used for the training of MetaBBO methods and the rest 25\% instances are taken as testing set. The ``difficult'' postfix on the contrary uses 25\% for training and 75\% for testing. The maximal function evaluations of all baselines are 20,000 for Synthetic and Noisy Synthetic Testsuites, 1,000 for Protein-Docking Testsuite. All baselines are trained for $1.5\times10^6$ learning steps and tested over 51 independent runs to ensure fairness. Other baseline hyper-parameters follow the settings in MetaBox. The experiments are conducted on Intel(R) Xeon(R) Gold 6254 CPU with 64GB RAM and NVIDIA GeForce RTX 4090 GPU. More details of train-test instance split and baseline settings can be found in the MetaBox repository. We hope this response could clear the reviewers' concerns.

\section{Additional Experimental Results}
\subsection{AEI Results}\
For evaluating more algorithms, we could have evaluated the AEI scores of additional traditional algorithms covering DE, PSO, GA and ES on Synthetic and Protein-docking testsuites in Table~\ref{tab:tradition} below, as well as the AEI of MetaBBO methods in Table~\ref{tab:metabbo}. 
\begin{table}[t]
\centering
\caption{The AEI scores of traditional algorithms on Synthetic and Protein-docking testsuites. The underlined algorithms are the algorithms presented in the paper.}
\label{tab:tradition}
\resizebox{\columnwidth}{!}{%
\begin{tabular}{c|cccccccccccccccc}
\hline
 & DE & \underline{JADE} & SHADE & MadDE & AMCDE & PSO & FIPSO & sDMSPSO & \underline{GLPSO} & EPSO & GA & AGA & ES & \underline{CMA-ES} & Sep-CMA-ES & IPOP-CMA-ES \\ \hline
\begin{tabular}[c]{@{}c@{}}Synthetic\end{tabular} &
  16.10 &
  13.23 &
  14.11 &
  12.97 &
  12.88 &
  14.13 &
  12.93 &
  9.25 &
  13.40 &
  13.17 &
  12.86 &
  13.61 &
  14.26 &
  22.61 &
  23.31 &
  22.22 \\
\begin{tabular}[c]{@{}c@{}}easy\end{tabular} &
  $\pm$1.68 &
  $\pm$1.46 &
  $\pm$1.54 &
  $\pm$1.35 &
  $\pm$1.39 &
  $\pm$1.45 &
  $\pm$1.52 &
  $\pm$1.25 &
  $\pm$1.08 &
  $\pm$1.31 &
  $\pm$1.38 &
  $\pm$1.44 &
  $\pm$1.43 &
  $\pm$4.62 &
  $\pm$4.98 &
  $\pm$4.59 \\ \hline
\begin{tabular}[c]{@{}c@{}}Synthetic\end{tabular} &
  9.45 &
  9.90 &
  11.24 &
  9.70 &
  9.83 &
  7.21 &
  9.69 &
  6.77 &
  10.22 &
  10.01 &
  9.22 &
  11.65 &
  15.23 &
  15.98 &
  16.50 &
  16.24 \\
\begin{tabular}[c]{@{}c@{}}difficult\end{tabular} &
  $\pm$1.12 &
  $\pm$1.19 &
  $\pm$1.42 &
  $\pm$1.25 &
  $\pm$1.12 &
  $\pm$0.95 &
  $\pm$1.24 &
  $\pm$0.89 &
  $\pm$1.15 &
  $\pm$1.14 &
  $\pm$1.06 &
  $\pm$1.20 &
  $\pm$1.36 &
  $\pm$2.88 &
  $\pm$2.29 &
  $\pm$2.58 \\ \hline
\begin{tabular}[c]{@{}c@{}}Protein\end{tabular} &
  0.82 &
  0.95 &
  0.96 &
  1.00 &
  0.94 &
  0.84 &
  0.87 &
  1.13 &
  1.19 &
  1.09 &
  1.08 &
  1.13 &
  0.90 &
  0.93 &
  0.94 &
  0.94 \\ 
\begin{tabular}[c]{@{}c@{}}easy\end{tabular} &
  $\pm$0.03 &
  $\pm$0.03 &
  $\pm$0.03 &
  $\pm$0.04 &
  $\pm$0.02 &
  $\pm$0.03 &
  $\pm$0.02 &
  $\pm$0.04 &
  $\pm$0.04 &
  $\pm$0.03 &
  $\pm$0.02 &
  $\pm$0.03 &
  $\pm$0.03 &
  $\pm$0.04 &
  $\pm$0.04 &
  $\pm$0.04 \\ \hline
\begin{tabular}[c]{@{}c@{}}Protein\end{tabular} &
  0.76 &
  0.93 &
  0.89 &
  0.86 &
  0.86 &
  0.84 &
  0.86 &
  0.85 &
  1.05 &
  1.02 &
  1.02 &
  1.09 &
  0.88 &
  0.92 &
  0.94 &
  0.93 \\
\begin{tabular}[c]{@{}c@{}}difficult\end{tabular} &
  $\pm$0.02 &
  $\pm$0.03 &
  $\pm$0.03 &
  $\pm$0.03 &
  $\pm$0.04 &
  $\pm$0.04 &
  $\pm$0.02 &
  $\pm$0.02 &
  $\pm$0.03 &
  $\pm$0.03 &
  $\pm$0.03 &
  $\pm$0.02 &
  $\pm$0.04 &
  $\pm$0.04 &
  $\pm$0.04 &
  $\pm$0.04 \\ \hline
\end{tabular}%
}
\end{table}

\begin{table}[t]
\centering
\caption{The AEI scores of MetaBBO algorithms on Synthetic and Protein-docking testsuites. }
\label{tab:metabbo}
\resizebox{0.65\columnwidth}{!}{%
\begin{tabular}{c|cccccccccc}
\hline
 & RL-DAS & DE-DDQN & LDE & RLEPSO & LES & GLEET & SYMBOL & RNN-OI & GLHF & OPRO \\ \hline
\begin{tabular}[c]{@{}c@{}}Synthetic\end{tabular} &
  14.29 &
  8.97 &
  12.06 &
  16.09 &
  2.92 &
  5.43 &
  12.67 &
  1.40 &
  7.53 &
  1.72 \\ 
\begin{tabular}[c]{@{}c@{}}easy\end{tabular} &
  $\pm$1.45 &
  $\pm$0.98 &
  $\pm$1.21 &
  $\pm$2.43 &
  $\pm$0.98 &
  $\pm$1.08 &
  $\pm$1.08 &
  $\pm$0.13 &
  $\pm$2.81 &
  $\pm$0.73 \\ \hline
\begin{tabular}[c]{@{}c@{}}Synthetic\end{tabular} &
  10.86 &
  7.85 &
  9.84 &
  9.36 &
  2.16 &
  5.02 &
  12.02 &
  0.02 &
  7.12 &
  1.61 \\
\begin{tabular}[c]{@{}c@{}}difficult\end{tabular} &
  $\pm$1.19 &
  $\pm$0.82 &
  $\pm$1.09 &
  $\pm$1.39 &
  $\pm$0.82 &
  $\pm$1.15 &
  $\pm$1.15 &
  $\pm$0.01 &
  $\pm$1.38 &
  $\pm$1.33 \\ \hline
\begin{tabular}[c]{@{}c@{}}Protein\end{tabular} &
  0.99 &
  0.71 &
  0.98 &
  1.03 &
  1.11 &
  1.21 &
  0.99 &
  1.00 &
  1.13 &
  0.22 \\
\begin{tabular}[c]{@{}c@{}}easy\end{tabular} &
  $\pm$0.03 &
  $\pm$0.02 &
  $\pm$0.03 &
  $\pm$0.03 &
  $\pm$0.04 &
  $\pm$0.03 &
  $\pm$0.04 &
  $\pm$0.03 &
  $\pm$0.03 &
  $\pm$0.01 \\ \hline
\begin{tabular}[c]{@{}c@{}}Protein\end{tabular} &
  0.91 &
  0.70 &
  0.90 &
  1.02 &
  0.95 &
  1.01 &
  0.91 &
  0.98 &
  1.01 &
  0.22 \\
\begin{tabular}[c]{@{}c@{}}difficult\end{tabular} &
  $\pm$0.02 &
  $\pm$0.02 &
  $\pm$0.03 &
  $\pm$0.03 &
  $\pm$0.03 &
  $\pm$0.02 &
  $\pm$0.04 &
  $\pm$0.05 &
  $\pm$0.02 &
  $\pm$0.01 \\ \hline
\end{tabular}%
}
\end{table}

\subsection{Comparison on the Learning Capabilities}
As a learning based paradigm, MetaBBO should also be evaluated using metrics that reflect its learning effectiveness. Next, we evaluate the MetaBBO methods using the Meta Generalization Decay (MGD) and Meta Transfer Efficiency (MTE) from MetaBox. By presenting the scores of MGD and MTE of these MetaBBO methods, the learning true learning abilities of these MetaBBO methods could be in-depth discussed. We note that this learning ability testing procedure could be generalized to other MetaBBO scenarios beyond single-objective optimization. We here use single-objective testsuites in MetaBox as a showcase for our readers.

\subsubsection{Meta Generalization Decay}
MGD measures the generalization performance of a MetaBBO method for unseen tasks. Concretely, MetaBox trains two models for the MetaBBO method on two source suites ($A$ and $B$) and test them on the target suit $B$. We record the AEI scores of these two models on the target suit as $AEI_A$ and $AEI_B$ respectively. The MGD$(A,B)$ is computed as
\begin{equation}
    \text{MGD}(A,B) = 100\times(1-\frac{AEI_A}{AEI_B})\%,
\end{equation}
A smaller MGD score indicates that the method generalizes well from $A$ to $B$. 

\begin{figure*}[ht]
    \centering
    \includegraphics[width=\linewidth]{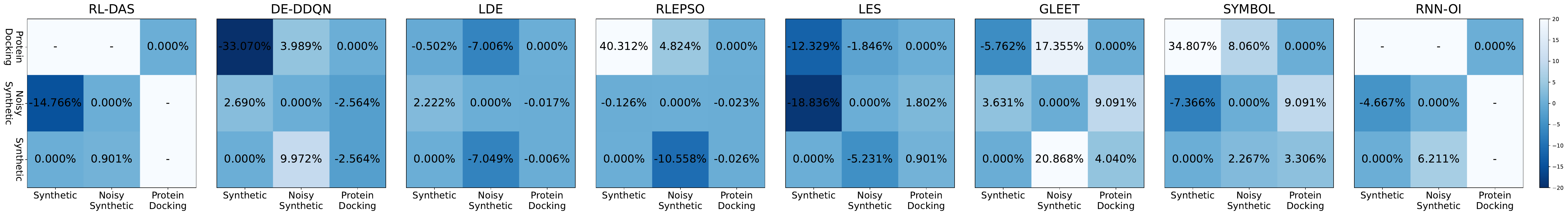}
    \caption{MGD scores of baselines. The value at $i$-th row and $j$-th column is the MGD$(i,j)$, with smaller value indicating better performance.}
    \label{fig:mgd}
\end{figure*}

Fig.~\ref{fig:mgd} shows the MGD plot of MetaBBO baselines, with GLHF omitted since it does not provide training codes. The `-' indicates that the model fails to generalize to target testsuit. We can observe that: 1)~RL-DAS and RNN-OI cannot be generalized from Synthetic-10D to Protein Docking-12D due to their optimization state features are dimension-dependent, highlighting the importance of optimization state design. 
2)~MetaBBO-NE methods~(LES) achieves more robust generalization than MetaBBO-RL~(RL-DAS, DE-DDQN, LDE, RLEPSO, GLEET, SYMBOL) and MetaBBO-SL~(RNN-OI) baselines, possible revealing the learning effectiveness advantage of neuroevolution paradigm due to its global learning ability. However, neuroevolution is exponentially resource-consuming as the neuron counts scale, implying a tradeoff between effectiveness and efficiency.
3)~Larger models~(e.g., GLEET with 3 Transformer layers) underperform smaller ones~(e.g., DE-DDQN with a single MLP) in the generalization evaluation, even though they outperform within the training distribution. Since the generalization performance is closely tied to the model capacity and the data scale, further investigation into the scaling laws in MetaBBO is highly anticipated.

\begin{table}[t]
\centering
\caption{MTE scores of the MetaBBO methods in the transfer from source testsuites to target testsuites.}
\label{tab:MTE}
\resizebox{0.5\columnwidth}{!}{%
\begin{tabular}{c|c|c|c|c|c|c}
\hline
  \begin{tabular}[c]{@{}c@{}}Source\\Testsuites\end{tabular} &
  \multicolumn{2}{c|}{Synthetic} &
  \multicolumn{2}{c|}{\begin{tabular}[c]{@{}c@{}}Noisy-Synthetic\end{tabular}} &
  \multicolumn{2}{c}{\begin{tabular}[c]{@{}c@{}}Protein-Docking\end{tabular}}
   \\ \hline
  \begin{tabular}[c]{@{}c@{}}Target\\Testsuites\end{tabular} &
  \begin{tabular}[c]{@{}c@{}}Noisy\\ Synthetic\end{tabular} &
  \begin{tabular}[c]{@{}c@{}}Protein\\ Docking\end{tabular} &
  Synthetic &
  \begin{tabular}[c]{@{}c@{}}Protein\\ Docking\end{tabular} &
  Synthetic &
  \begin{tabular}[c]{@{}c@{}}Noisy\\ Synthetic\end{tabular} \\ \hline
RL-DAS  & fail   &  -  & 1  & -  &  - &  - \\ \hline
DE-DDQN &  -9.5  &  1  & -3.2  &  fail &-1.41   & -9.5  \\ \hline
LDE     & -0.05 & -0.17  & 0.19  & -0.17  & 1  & -0.05  \\ \hline
RLEPSO  & fail   & -4.25  & -0.39 & -4.25  & 1  & fail  \\ \hline
LES     &  -1.65  & -0.05   & fail  & -0.05  & fail  & 1  \\ \hline
GLEET   & 1  &  -0.05  & fail  & -0.05  & fail  & -0.39  \\ \hline
SYMBOL  & -0.16   & fail   & 0.97  & -2.17  & -0.05  & fail  \\ \hline
RNN-OI  &  1  &  -  & 0.01  & -  & -  &  - \\ \hline
\end{tabular}%
}
\end{table}

\subsubsection{Meta Transfer Efficiency}
MTE score measures the transfer learning capability. For a MetaBBO method, its MTE from a problem set $A$ to $B$ is computed as: 
\begin{equation}
    \text{MTE}(A,B) = 100\times(1-\frac{T_{\mathrm{finetune}}}{T_{\mathrm{scratch}}})\%,
\end{equation} 
where $T_{\mathrm{scratch}}$ is the learning steps used to attain best performance when training on $B$. $T_{\mathrm{finetune}}$ is the learning steps used to fine-tune a model trained on $A$ to attain the same performance level.  
A larger MTE score indicates that the knowledge learned in $A$ can be easily transferred to solve $B$. 
Table~\ref{tab:MTE} presents the MTE scores of all baselines under each pair of source-target problem collections, where `fail' indicates the baseline can not be fine-tuned to achieve similar performance level on the target problems.
Results show that: 1)~While many baselines highlight their transfer learning ability under some cases~(e.g., GLEET: Synthetic to Noisy-Synthetic), they show transfer limitations in other cases, suggesting room for improvement. 2)~The overall transferring performances across all baselines and problem collections are relatively noisy, making it difficult to determine whether some transfer failures stem from the algorithm designs or the diversity of the problem collections. This opens up a research opportunity to explore the relationship between problem diversity and generalization, as well as how to construct ``good'' training set for MetaBBO.

\end{document}